\pdfoutput=1

\documentclass[11pt]{article}

\usepackage[final]{acl}
\usepackage{multirow}
\usepackage{xcolor}
\usepackage{color, colortbl}
\usepackage{rotating}

\definecolor{mygray}{gray}{0.9}
\usepackage{booktabs}
\usepackage{amsmath} 
\usepackage{amssymb}
\usepackage{listings}

\DeclareMathOperator*{\argmax}{arg\,max}
\usepackage{algorithm}
\usepackage{algpseudocode}
\usepackage{times}
\usepackage{latexsym}

\usepackage[T1]{fontenc}

\usepackage[utf8]{inputenc}

\usepackage{microtype}

\usepackage{inconsolata}
\usepackage{subcaption}  

\usepackage{graphicx}

\title{Attributes as Textual Genes: Leveraging LLMs as Genetic Algorithm Simulators for Conditional Synthetic Data Generation}

\author{Guangzeng Han, Weisi Liu, Xiaolei Huang \\
  Department of Computer Science, University of Memphis \\
  Memphis, TN, USA\\
  \texttt{\{ghan, wliu9, xiaolei.huang\}@memphis.edu}}

\begin{document}
\maketitle
\begin{abstract}

Large Language Models (LLMs) excel at generating synthetic data, but ensuring its quality and diversity remains challenging. 
We propose \textbf{Genetic Prompt}\footnote{The code is available at: \url{https://github.com/trust-nlp/Genetic-Prompt}}, a novel framework that combines genetic algorithms with LLMs to augment synthetic data generation. 
Our approach treats semantic text attributes as gene sequences and leverages the LLM to simulate crossover and mutation operations. 
This genetic process enhances data quality and diversity by creating novel attribute combinations, yielding synthetic distributions closer to real-world data. 
To optimize parent selection, we also integrate an active learning scheme that expands the offspring search space. 
Our experiments on multiple NLP tasks reveal several key findings: Genetic Prompt not only significantly outperforms state-of-the-art baselines but also shows robust performance across various generator model sizes and scales. 
Moreover, we demonstrate that fusing our synthetic data with the original training set significantly boosts downstream model performance, particularly for class-imbalanced scenarios. 
Our findings validate that Genetic Prompt is an effective method for producing high-quality synthetic data for a wide range of NLP applications.

\end{abstract}

\section{Introduction}
\label{sec:intro}

Leveraging large language models (LLMs) to synthesize training data for smaller models is a promising path to efficiency and accuracy across NLP tasks~\cite{long2024llms}, yet improving the quality and diversity of such data remains underexplored~\cite{ye2022zerogen,yue2023large}.
Prior work often steers synthetic data generation with pre-defined conditions such as clinical named entities~\cite{xu2024knowledge,josifoski2023exploiting}, label correlations~\cite{chung2023increasing}, and writing styles~\cite{xu2024knowledge,yue2023large}. However, reliance on static cues limits model reasoning, cross domain generalization, and dataset diversity. Recent findings further suggest that instructional diversity is crucial for high quality synthesis~\cite{kim2025evaluating, xu2024knowledge, li2024entropic}, which motivates our core question: \textit{how can we automatically amplify data diversity and generator adaptability for robust downstream training?}

Inspired by the genetic algorithms~\cite{mitchell1980need} (GAs), we propose \textbf{Genetic Prompt} framework for generating high-quality synthetic data that 1) automatically identifies and treats textual attributes as gene sequences, 2) integrates active learning to optimize parent selection process, and 3) diversifies synthetic data by simulating the genetic processes of crossover and mutation.

GAs are population-based search heuristics inspired by the process of natural selection iteratively evolving solutions to optimization problems~\cite{mitchell1980need}.
However, optimizing synthetic data generation via genetic method and LLMs has rarely been explored, due to two primary issues.
First, assessing the fitness of individual samples is hard for downstream models, because a data point’s effect is context-dependent and hinges on high-order interactions with the rest of the training set, as shown by recent study~\cite{ilyas2022datamodels}.
Second, data characteristics can be diverse and complex, while only focusing on partial scopes of target data may not be applicable to yield a large amount of high-quality synthetic data by LLMs.
For example, a close study~\cite{liu2024autodan} alters words and sentences on the syntactic level to continue optimizing the best individual samples.
This approach is inherently constrained to generating only a small volume of data, limiting its scalability for large-scale synthetic data generation. 



To mitigate the unreliability of per example fitness in selecting parents, we replace conventional fitness based selection with an active learning (AL) strategy~\cite{zhang2023llmaaa, Ren2021survey-AL}. 
In each round, AL pairs previously unused, diverse parents, enlarging the search space for offspring generation. 
To improve coverage and scalability, we avoid operating on individual tokens or sentences or treating them as genes. 
Instead, we encode higher level linguistic and semantic attributes (e.g., readability and sentence structure) as genes. This design lets us leverage the knowledge and comprehension abilities of LLMs to implement crossover and mutation at the attribute level, producing large volumes of high quality and diverse data.

Our experiments demonstrate that Genetic Prompt consistently outperforms state-of-the-art baselines like AttrPrompt~\cite{yue2023large} and Curated LLM~\cite{seedat2024curated}, while also exhibiting superior robustness and scalability across various generator LLM sizes. 
Our qualitative analysis shows that the data generated by Genetic Prompt has superior semantic and lexical diversity and more accurately captures the statistical characteristics of real data. 
Furthermore, we find that incorporating this synthetic data with gold data can significantly boost model performance, particularly in data imbalance settings. 
These findings validate the effectiveness of our Genetic Prompt framework for producing high-quality synthetic data for downstream tasks.

\section{Genetic Prompt for Synthetic Data Generation}
\label{sec:method}

\begin{figure*}[tbph]
    \centering
    \includegraphics[width=.99\textwidth]{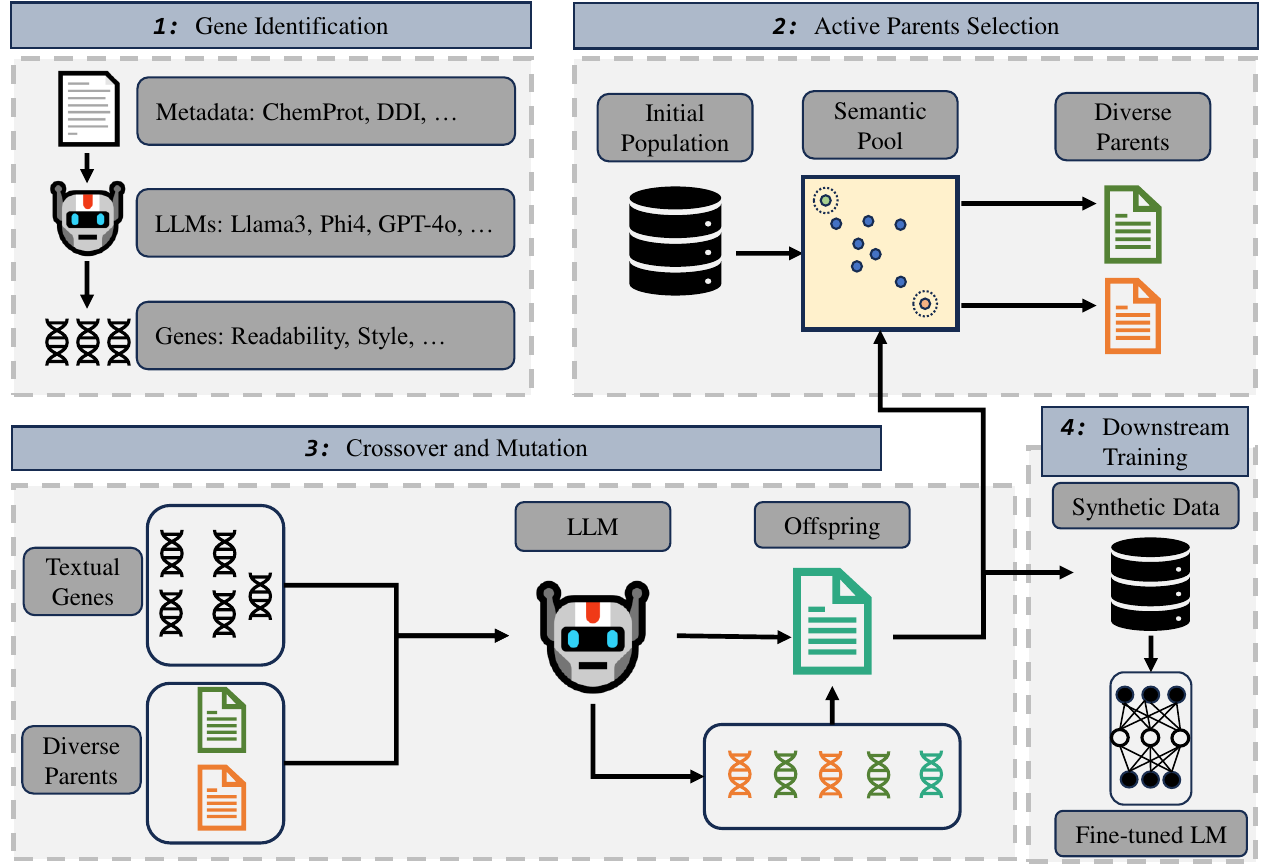}
    \caption{Overview of our Genetic Prompt framework. }
    \label{fig:framework}
\end{figure*}

In this section,  we present our Genetic Prompt framework in Figure~\ref{fig:framework}.
The key idea is to automatically identify and treat data attributes as genes and prompt LLMs to do crossover and mutation operations over a small initial population from gold data.
It consists of four major modules: 1) textual genes identification, 2) active parents selection, 3) crossover and mutation, and 4) downstream training. 
We include detailed algorithmic process in Algorithm~\ref{alg:LLM-GA} under the Appendix.

\subsection{Textual Genes Identification}

To improve the quality of synthetic data or diversify its style, attribute-aware conditional text generation a promising direction. 
However, these methods generally rely on attribute values strictly predefined by humans~\cite{yu2021attribute,russo2020control,logeswaran2018content} or LLMs~\cite{xu2024knowledge,yue2023large}. 
In contrast, our approach views data attributes as ``genes'' of the text. 
With just a single round of interaction with LLMs, we identify the genes without predefining specific values for each gene, avoiding excessive constraints on the LLMs and reducing bias from human selection.
Specifically, our solution involves inputting the metadata (such as ``task type: chemical protein interaction'') and task samples into LLMs and posing questions like ``If the attributes of the text are viewed as genes, which genes are most important in the given chemical protein interaction extraction task?'' 
Given the Chemprot~\cite{chemprot} data as an example, we received responses including \textit{length, sentence structure, entity proximity, polarity}, etc. 
Subsequently, following previous work~\cite{yue2023large}, we adopt a human-AI collaboration scheme to empirically select vital genes (examples in Table~\ref{tab:textual_genes} under Appendix~\ref{sec: selected genes}).
The textual genes identification process can be formalized as: 
$$G = \Phi(M, S, \textit{Ins.}_{G})$$,
where $G$ represents the set of identified textual genes, $\Phi$ represents the extraction function, $M$ is the task metadata, $S$ is the set of initial samples, and $\textit{Ins.}_{G}$ is the prompt template used to guide the extraction process. These genes serve as the fundamental units for our subsequent genetic operations.

\subsection{Active Parents Selection}

In genetic algorithms, the primary objective is to select the fittest pair of parents $(p_i^*, p_j^*)$ for crossover and mutation, thereby producing offspring with enhanced characteristics. 
However, in synthetic data generation, evaluating fitness of sample pairs is challenging. 
Active learning, which assumes that samples vary in importance and contribute differently to performance improvements, offers a promising alternative. 
The standard pool-based active learning paradigm~\cite{Ren2021survey-AL,zhang2023llmaaa} is designed to extract the most informative and diverse samples from an unlabeled pool to enhance model generalization. 
Inspired by these insights, we adopt an active learning strategy for parent selection in our framework. 
Specifically, we extract a previously unused pair of samples from a data pool $\{ P, E\}$, where $P=\{p_0 , p_1, \dots , p_n \}$ is the current population of size $n$
and $E=\{e_0 , e_1, \dots , e_n \}$ denotes their semantic representations~\footnote{We use the Sentence Transformer~\cite{reimers2019sentence} to encode textual populations and obtain their semantic representations.}. 
The extracted pair of samples is the one that exhibits the largest semantic distance, thereby maximizing the offspring search space during crossover and mutation.
A larger semantic distance indicates that the selected samples differ significantly in meaning, increasing the likelihood of generating more diverse offspring when combined. 
To quantify this, we compute the Euclidean distance between their semantic representations and choose the pair with the greatest distance, thus broadening the search space during the evolutionary process.
The parents selection process can be formulated as:
$$(p_i^*, p_j^*) = \arg\max_{e_i, e_j \in E} \text{dist}(e_i, e_j)$$

\begin{table*}[ht]
\centering
\resizebox{0.98\textwidth}{!}{%
\begin{tabular}{l|cccccccc}
\toprule
Data& \textbf{AGnews} 
  & \textbf{StackExchange} 
  & \textbf{Chemprot} 
  & \textbf{DDI} 
  & \textbf{Semeval} 
  & \textbf{Conll04} 
  & \textbf{SciTLDR} 
  & \textbf{MeQSum} \\ 
\hline
Train                   
      &  120,000    & 27,086  & 1,069 & 1,482 & 6,590 & 922   & 2168    & 1,000     \\
Test                    
       & 7,600     & 2,494    & 1,041 & 339   & 2,263 & 288   & 662     & 100     \\
Class                   
      & 4     & 50     & 5     & 4     & 9     & 5     & -     & -    \\
Domain                  
  & News & Science & Biomedicine & Pharmacology & Web & News & Science & Medical \\

Task               
  & CLS & CLS & RE & RE & RE & RE & ABS & ABS \\
Imbalance Ratio &1.00 &1,283.77 &7.16 &  8.08 &1.99 &1.51 &-&- \\
\bottomrule
\end{tabular}}
\caption{Statistics of eight datasets from three different NLP tasks with varying data sizes and domains. CLS denotes text classification, RE refers to relation extraction, and ABS means text summarization. Imbalance Ratio represents the ratio of largest to smallest class size.}
\label{tab:data}
\end{table*}

\subsection{Crossover and Mutation}

We denote the genetic algorithm prompts for class $l$ as $\textit{Ins}^l_{\textit{GA}}$ and refer $\rho(\cdot)$ to the generation process from LLMs, where the distribution of generated text is conditioned on the prompt and other input parameters. 
In each iteration, we leverage an LLM as an evolution simulator, performing crossover and mutation processes to generate new textual offsprings.
The overall generation process can be formalized as:
$$\mathcal{D}_{\text{syn}} \sim \rho_{l\sim \mathcal{Y}}(\cdot; \text{Ins}^l_{\text{GA}}, (p_i^*, p_j^*), G)$$
, where $\text{Ins}^l_{\text{GA}}$ is the genetic algorithm prompt template for label $l$, $(p_i^*, p_j^*)$ are the selected parent samples, $G$ represents the identified textual genes.
Specifically, we utilize the two selected parent examples $(p_i^*, p_j^*)$ from the current population and perform the \textit{crossover and mutation process} until we reach the target population size $N$.

\paragraph{Crossover:}
To fully utilize the diverse ``genes'' (e.g., paragraph structure, entity ordering, semantic stance) embedded in the parent texts, we randomly sample and blend key elements from both parents to form a hybrid offspring. 
We randomly partition the identified textual genes $G$ into three groups: $G = G_1 \cup G_2 \cup G_3$, where genes in $G_1$ are inherited from parent $p_i^*$, genes in $G_2$ are inherited from parent $p_j^*$, and genes in $G_3$ will undergo mutation.
This step ensures that the offspring text preserves the primary content of its parents while organically combining their structure and expressions.

\begin{itemize}
    \item \textbf{Parent 1:} 
    ``We investigated the synergistic effects of \textit{Compound A} on \textit{Protein B}. Preliminary results suggest partial agonistic activity.''
    \item \textbf{Parent 2:} 
    ``Our study shows that \textit{Drug X} significantly activates \textit{Receptor Y} in neuronal cells, indicating a potent agonist relationship.''
    \item \textbf{Offspring (via Crossover):} 
    ``We investigated how \textit{Drug X} interacts with \textit{Receptor Y} in neuronal cells. Preliminary results suggest partial agonistic activity.''
\end{itemize}

In this illustration, \textit{Drug X} and \textit{Receptor Y} are inherited from Parent 2, while the sentence structure and conclusion are drawn from Parent 1. 
By inheriting crucial ``genes'' from parents, the newly generated text retains essential content and has the potential to introduce new textual variations, broadening coverage and diversity in the dataset.

\paragraph{Mutation:}

Unlike token- or sentence-level modification (e.g., swapping words or paraphrasing), our mutation process targets the \emph{semantic-level} identified in Section~\ref{sec: selected genes}. 
Specifically, we randomly alter the genes in $G_3$ that were not inherited during crossover—such as positioning or relationship between entities, polarity or sentiment of a statement, or text’s functional role—while keeping the overall meaning relevant to the task. 
The mutated offspring equip more diverse semantic attributes than simple lexical substitutions and thus can diverge more meaningfully from the parent samples with broaden the evolutionary search space.

\subsection{Downstream Training}

We apply synthetic data $D_{\text{Syn}}$ for downstream tasks using pre-trained language models and cross-entropy as the loss function. 
To validate the effectiveness of synthetic datasets, we follow a standard fine-tuning process by a uniform learning rate to both the pre-trained and prediction layers without any warm-up or decay.

\section{Experimental Settings}
\label{sec:experimental settings}
\subsection{Datasets}

We use eight publicly available datasets: AGNews~\cite{agnews},  StackExchange~\cite{geigle2021tweac},  Chemprot~\cite{chemprot}, DDI~\cite{DDI}, Semeval2010~\cite{SEMEVAL2010}, Conll04~\cite{conll04}, SciTLDR~\cite{cachola2020tldr} and MeQSum~\cite{ben2019summarization-MeQSum}. 
We summarize data statistics in Table~\ref{tab:data} with more details of these benchmarks in Appendix~\ref{appendix:data}.

\subsection{Technical Details}

We experiment with open-sourced (e.g., \texttt{Phi4} \cite{abdin2024phi} and \texttt{Llama3.1-70B} \cite{dubey2024llama}) and proprietary LLMs (\texttt{GPT-3.5-turbo} and \texttt{GPT-4o}~\cite{openai2022chatgpt}) as our data generators.
And we set the temperature and p-value of all LLMs to 1 to ensure reproducibility.
Our empirical analysis in Sec~\ref{sec:Empirical Analysis} uses multiple LLMs (\texttt{Llama3.2-3B}, \texttt{Llama3.1-8B}, and \texttt{Llama3.1-70B}) trained on the same data source.
For each dataset, we generate the same number of samples across different methods and ablation settings.
Unless specified otherwise, \texttt{Llama3.1-70B} is our default data generator, and \texttt{RoBERTa-base}~\cite{liu2019roberta} and \texttt{T5-large}~\cite{raffel2020exploring} are downstream models for classification and summarization, respectively.

\subsection{Baselines}
Since our Genetic Prompt requires a small amount of real data to initialize the population, for a fair comparison, we compare the genetic prompt against three approaches that can be integrated with the few-shot setting: 1) \textbf{SimPrompt}~\cite{ye2022zerogen,yue2023large}, 2) \textbf{AttrPrompt}~\cite{yue2023large} and 3) \textbf{Curated LLM (CLLM)}~\cite{seedat2024curated}.
To keep consistence, all the baselines are also provided with two examples randomly selected from the same pool of real samples used for initializing our Genetic Prompt.
For \textit{SimPrompt}, we use its original class-conditional setting, where prompts guide text generation based on predefined categories. SimPrompt follows a template-based approach, such as: \textit{The \{class\_name\} relation means \{label\_def\}, Your task is to write a sentence about \{class\_name\} relation between 2 entities.}
Since \textbf{AttrPrompt} is a close work using textual attributes, we ensure consistency by adopting the textual genes used in Genetic Prompt as its required attributes. 
Following original settings of \textit{AttrPrompt}, we then prompt \textit{GPT-4-Turbo} to generate and select specific attribute values and then form various unique combinations. 
For example, given the possible values for each attribute in Conll04 data (e.g., five different writing styles and five different domains), the total number of unique combinations exceeds 4,000. 
For the \textit{Curated LLM (CLLM)} approach, we apply the proposed learning dynamics method to curate the over-generated dataset by \textit{Simprompt}.

\begin{table}[htp]
\setlength{\tabcolsep}{6pt}

\resizebox{0.485\textwidth}{!}{
\begin{tabular}{ll|ccccc}
\toprule
\textbf{Data} & \textbf{Method} 
  & \textbf{APS} $\downarrow$ 
  & \textbf{Intra APS} $\downarrow$ 
  & \textbf{Inter APS} $\downarrow$ 
  & \textbf{CMD} $\downarrow$ 
  & \textbf{Vocab.} \\
\hline

\multirow{5}{*}{AGnews} 
  & SimPrompt    & 0.167 & 0.325 & 0.113 & 0.853 & \textbf{33,249} \\
  & AttrPrompt   & 0.205 & 0.294 & 0.174 & 1.073 & 29,049 \\
  & Curated LLM  & 0.250 & 0.324 & 0.224 & 1.203 & 31,051 \\
  & Ours         & \textbf{0.129} &\textbf{ 0.260} & \textbf{0.084} & \textbf{0.716} & 24,459 \\
  & Gold         & 0.029 & 0.076 &0.013 & -    & 124,673 \\
\hline
\multirow{5}{*}{StackExchange} 
  & SimPrompt    & 0.377 & 0.611 & 0.372 & 0.706 & \textbf{127,595} \\
  & AttrPrompt   & \textbf{0.306} & 0.576 &\textbf{ 0.300} & 0.741 & 46,791 \\
  & Curated LLM  & 0.323 & \textbf{0.488} & 0.319 & \textbf{0.520} & 119,482 \\
  & Ours         & 0.324 & 0.491 & 0.321 & 0.595 & 81,948 \\
  & Gold         & 0.231 & 0.256 & 0.225 & -     & 106,967 \\
\hline

\multirow{5}{*}{Chemprot} 
    & SimPrompt    & 0.423 & 0.660 & 0.364 & 0.819 & 1,847 \\
    & AttrPrompt   & 0.432 & \textbf{0.461} & 0.425 & 0.720  & 2,115\\
    & Curated LLM  & 0.425 & 0.658 & 0.366 & 0.819  & 1,814 \\
    & Ours         & \textbf{0.389} & 0.510 & \textbf{0.359} & \textbf{0.628}  &\textbf{4,596}\\
    & Gold         & 0.326 & 0.352 & 0.314 & - & 4,123\\
\hline
\multirow{5}{*}{DDI} 
    & SimPrompt    & 0.572 & 0.691 & 0.532 & 1.077& 1,240 \\
    & AttrPrompt   & 0.516 & 0.542 & 0.508 & 0.925 &1,556 \\
    & Curated LLM  & 0.569 & 0.689 & 0.530 & 1.065  &1,263\\
    & Ours         & \textbf{0.456} & \textbf{0.526} & \textbf{0.432} &\textbf{0.706}  &\textbf{3,524}\\
    & Gold         & 0.397 & 0.410 & 0.390 &  - & 3,472\\
\hline
\multirow{5}{*}{Semeval2010} 
    & SimPrompt    & 0.407 & 0.618 & 0.380 & 1.188  & 3,093\\
    & AttrPrompt   & 0.451 & 0.499 & \textbf{0.445} & 1.173 & 5,174\\
    & Curated LLM  & 0.405 & 0.617 & 0.380 & 1.188 & 3,049\\
    & Ours         & \textbf{0.398} &\textbf{ 0.460} & 0.390 & \textbf{1.093} &\textbf{9,425} \\
    & Gold         & 0.248 & 0.280 & 0.244 &  - & 19,268\\
\hline
\multirow{5}{*}{Conll04} 
    & SimPrompt    & 0.352 & 0.587 & 0.239 & 1.140 & 2,683 \\
    & AttrPrompt   & 0.379 & 0.444 & 0.362 & 0.973 &3,850 \\
    & Curated LLM  & 0.358 & 0.599 & 0.297 & 1.163 & 2,653\\
    & Ours         & \textbf{0.292} & \textbf{0.385} & \textbf{0.269} &\textbf{0.852} & \textbf{8,381}\\
    & Gold         & 0.177 & 0.231 & 0.162 &  - &6,563\\
\hline

\multirow{5}{*}{SciTLDR} 
  & SimPrompt    & 0.746 & - & - & 3.600 & 2,491 \\
  & AttrPrompt   & \textbf{0.375} & - & - &\textbf{2.436} &5,527 \\
  & Curated LLM  & - & - & - & - & - \\
  & Ours         & 0.440 & - & - & 0.499 & \textbf{11,879} \\
  & Gold         & 0.386 & - & - & -     & 17,213 \\
\hline
\multirow{5}{*}{MeQSum} 
  & SimPrompt    & 0.316 & - & - &0.798 &4,514 \\
  & AttrPrompt   & \textbf{0.296} & - & -& \textbf{0.768} & 5,211 \\
  & Curated LLM  & - & - &- & -& - \\
  & Ours         & 0.316 &- &- & 0.834 & \textbf{7,268} \\
  & Gold         &0.176  &- & - & -     &8,483  \\
  \bottomrule
\end{tabular}}

\caption{Intrinsic analysis of synthetic data generation. $\downarrow$ indicates that lower values are better. We \textbf{Bold} the best scores except for ``Gold'', the original data. }
\label{tab:diversity}
\end{table}

\begin{table*}[ht]
\centering
\resizebox{\textwidth}{!}{%
\begin{tabular}{ll|ccccccccc}
\toprule
\multirow{2}{*}{Method} & \multirow{2}{*}{Model}

    & \multicolumn{1}{c}{AGnews} 
    & \multicolumn{1}{c}{StackExchange} 
    & \multicolumn{1}{c}{Chemprot}
    & \multicolumn{1}{c}{DDI}
    & \multicolumn{1}{c}{Semeval}
    & \multicolumn{1}{c}{Conll04}
    & \multicolumn{1}{c}{SciTLDR}
    & \multicolumn{1}{c}{MeQSum} \\
\cmidrule(lr){3-3}\cmidrule(lr){4-4}\cmidrule(lr){5-5}%
\cmidrule(lr){6-6}\cmidrule(lr){7-7}\cmidrule(lr){8-8}%
\cmidrule(lr){9-9}\cmidrule(lr){10-10}\cmidrule(lr){11-11}
 & & Micro-F1& Micro-F1
    & Micro-F1  & Micro-F1  & Micro-F1  & Micro-F1  
    & Rouge-L & Rouge-L \\
\hline
\multirow{4}{*}{SimPrompt}
 & Phi4           & 78.9$_{\;0.6}$  & 48.1$_{\;0.4}$  
                 & 65.9$_{3.7}$  & 49.1$_{1.5}$  & 58.0$_{1.3}$  & 65.0$_{2.0}$  
                 & 25.5$_{\;0.0}$  & 23.4$_{\;0.3}$ \\
 & Llama3.1-70B      & 82.3$_{\;1.9}$  & 51.8$_{\;0.3}$  
                 & 53.5$_{4.2}$  & 57.5$_{1.3}$  & 56.0$_{1.8}$  & 54.5$_{4.5}$  
                 & 26.0$_{\;0.1}$  & 22.6$_{\;0.2}$ \\
 & GPT-3.5-Turbo   & 71.8$_{\;0.3}$  & 40.2$_{\;0.1}$  
                 & 69.5$_{1.6}$  & 62.6$_{1.1}$  & 64.8$_{0.4}$  & 76.9$_{2.0}$  
                 & 25.7$_{\;0.3}$  & 23.5$_{\;0.2}$ \\
 & GPT-4o          & 81.8$_{\;1.0}$  & 52.2$_{\;1.5}$  
                 & 72.0$_{2.0}$  & 60.1$_{1.7}$  & 68.1$_{0.9}$  & 71.2$_{2.2}$  
                 & 26.0$_{\;0.2}$  & 23.0$_{\;0.2}$ \\
\hline
\multirow{4}{*}{AttrPrompt}
 & Phi4            & 79.4$_{\;2.0}$  & 50.3$_{\;0.0}$  
                 & 66.1$_{0.8}$  & 68.3$_{0.6}$  & 62.8$_{1.0}$  & 61.1$_{1.4}$  
                 & 26.3$_{\;0.0}$  & 25.2$_{\;0.1}$ \\
 & Llama3.1-70B      & 82.9$_{\;0.5}$  & 49.0$_{\;0.7}$  
                 & 66.0$_{1.4}$  & 63.8$_{0.9}$  & 67.1$_{0.8}$  & 66.4$_{1.6}$  
                 & 26.4$_{\;0.0}$  & 24.0$_{\;0.0}$ \\
 & GPT-3.5-Turbo   & 80.0$_{\;0.7}$  & 47.9$_{\;0.9}$  
                 & 68.1$_{2.5}$  & 62.1$_{5.0}$  & 71.0$_{0.8}$  & 68.8$_{2.3}$  
                 & 26.4$_{\;0.1}$  & 25.5$_{\;0.3}$ \\
 & GPT-4o         & 81.3$_{\;1.9}$  & 49.2$_{\;0.5}$  
                 & 73.6$_{2.5}$  & 61.9$_{2.8}$  & 71.1$_{0.8}$  & 73.3$_{3.6}$  
                 & 26.2$_{\;0.1}$  & 26.1$_{\;0.2}$ \\
\hline
\multirow{4}{*}{Curated LLM}
 & Phi4            & 79.7$_{\;0.6}$  & 45.5$_{\;1.2}$  
                 & 71.2$_{2.4}$  & 51.2$_{1.4}$  & 55.4$_{2.2}$  & 65.0$_{3.0}$  
                 & -  & - \\
 & Llama3.1-70B      & 82.5$_{\;1.0}$  & 50.0$_{\;0.4}$  
                 & 58.3$_{3.8}$  & 64.1$_{3.5}$  & 55.5$_{0.7}$  & 49.8$_{1.6}$  
                 & -  & - \\
 & GPT-3.5-Turbo   & 78.3$_{\;0.5}$  & 43.3$_{\;0.4}$  
                 & 71.7$_{0.9}$  & 63.8$_{2.6}$  & 65.1$_{0.9}$  & 73.3$_{1.4}$  
                 & -  & - \\
 & GPT-4o          & 82.3$_{\;1.4}$  & 50.4$_{\;0.6}$  
                 & 71.1$_{3.5}$  & 57.2$_{2.6}$  & 68.9$_{0.4}$  & 74.5$_{2.3}$  
                 & -  & - \\
\hline
\multirow{4}{*}{Ours}
 & Phi4            & 80.8$_{\;1.6}$  & \textbf{55.2}$_{\;0.4}$  
                 & 75.3$_{0.6}$  & 62.2$_{1.0}$  & 66.1$_{0.5}$  & \textbf{73.4$_{2.8}$}  
                 & 27.3$_{\;0.1}$  & \textbf{27.6}$_{\;0.1}$ \\
 & Llama3.1-70B      &\textbf{84.0}$_{\;1.3}$  & 51.2$_{\;0.9}$  
                 & \textbf{77.2$_{0.7}$} & \textbf{68.7$_{3.0}$} & \textbf{72.3$_{1.4}$} & 72.8$_{1.7}$  
                 & \textbf{27.5}$_{\;0.0}$  & 27.4$_{\;0.0}$ \\
 & GPT-3.5-Turbo   & 83.4$_{\;0.3}$  & 50.2$_{\;0.2}$  
                 & 77.9$_{2.1}$  & 68.7$_{2.0}$  & 73.2$_{0.8}$  & 67.5$_{3.4}$  
                 & 25.6$_{\;0.0}$  & 25.9$_{\;0.0}$ \\
 & GPT-4o          & \textbf{86.7}$_{\;0.8}$  & 49.8$_{\;0.4}$  
                 & \textbf{81.6$_{1.9}$} & \textbf{70.6$_{3.6}$} & \textbf{78.9$_{0.5}$} & \textbf{85.3$_{1.2}$}  
                 & \textbf{27.8}$_{\;0.1}$  & 27.2$_{\;0.3}$ \\
\bottomrule
\end{tabular}}
\caption{Extrinsic evaluation results on three downstream NLP tasks over 8 datasets. Each task was run three times randomly, and the subscripts are the standard deviations. We \textbf{bold} the best overall performance and the best performance among open-source LLMs.}

\label{tab:main results}
\end{table*}

\section{Results}
\label{sec:experimental results}
To evaluate the effectiveness of our Genetic Prompt method, we conducted four major studies, including intrinsic analysis of data quality, extrinsic evaluations on downstream tasks, integrative evaluation with real-world data, and ablation studies on individual modules. 
We also report extensive performance results and a case study in Appendix~\ref{sec:appendix-additional-results}.

\subsection{Intrinsic Analysis of Synthetic Data}
\label{sec:synthetic data analysis}
In an ideal scenario, synthetic data should faithfully reproduce the statistical characteristics of the real-world data it aims to emulate. 
Therefore, we employ multiple metrics to quantify the semantic diversity of synthetic data, as well as their distribution shift relative to the gold dataset.
These metrics include average pairwise sample similarity (APS), Central Moment Discrepancy (CMD)~\cite{zellinger2022central} and vocabulary size, which have detailed definition in Appendix~\ref{sec:definition of diversity}. 

We report intrinsic evaluation results in Table~\ref{tab:diversity}.
Among the synthetic generation methods, our Genetic Prompt stands out by achieving the best performance in APS, CMD, and vocabulary size, indicating that our approach can capture more semantic characteristics of the real-world data.
Notably, the gold data exhibits significantly lower APS scores compared to all synthetic datasets, which suggests that the inherent semantic diversity of real data is more salient.
On three datasets of Chemprot, DDI, and Conll04, the vocabulary size of our generated synthetic data even surpasses that of the gold data, indicating that our genetic framework may augment data semantic and lexical diversity  while closely mirror the distribution of real data.
We conduct a case study to examine the homogeneity issues in Appendix~\ref{sec:appendix-case-study}.

\subsection{Extrinsic Evaluation on NLP Tasks}
\label{subsec:main}
We conduct extrinsic evaluations on three standard NLP tasks on eight data with varying domains, such as biomedical and newspaper.
To examine generalizability, we evaluate multiple LLMs, adopt task specific metrics (e.g., F1 and Rouge scores) from the baselines~\cite{ye2022zerogen, yue2023large}, and report results in Table~\ref{tab:main results}.

Our Genetic Prompt consistently outperforms SimPrompt, AttrPrompt, and Curated LLM across major dataset.
In general, adopting GPT-4o as the data generator achieves better performance than other LLMs, which indicates a stronger LLM base is critical.
As the size of LLMs increases, the performance of our Genetic Prompt steadily improves, whereas other methods exhibit instability with occasionally significant declines (e.g., AttrPrompt plus GPT-4o drops by 6.4\% F1 score on DDI compared to Phi4). 
Lastly, while proprietary models generally outperform smaller open-source models, our Genetic Prompt using Phi4-14B model has either surpassed or closely matched the performance of baseline methods employing GPT-4o, which proves the effectiveness of our approach.

\subsection{Synthetic-Gold Data Fusion}

Will combining real-world and synthetic data improve downstream model performance?
To verify so, we merged the synthetic dataset with the original training set in equal proportions to train downstream models with detailed results in Table~\ref{tab:augmentation-combined}.

The fusion of synthetic and gold data demonstrates clear benefits for model performance across most tasks. 
Our Genetic Prompt consistently yields the greatest improvement among all baselines, achieving an average micro-F1 gain of 1.85\% and proving to be the only method that successfully benefits text summarization tasks (with +0.5 Rouge-L on SciTLDR and maintaining performance on MeQSum).

It is worth noting that our method shows particularly strong performance on class-imbalanced datasets, with significantly higher gains in macro-F1 compared to micro-F1 scores. 
On ChemProt, we achieve +3.2\% macro-F1 versus +2.3\% micro-F1; on DDI, +3.1\% macro-F1 versus +1.0\% micro-F1; and on StackExchange, +3.7\% macro-F1 versus +2.8\% micro-F1. 
This pattern indicates improved performance across all classes, especially benefiting minority classes.
In contrast, baseline methods often struggle with imbalanced datasets. SimPrompt shows negative performance on DDI (-0.6\% macro-F1, -1.1\% micro-F1), while AttrPrompt achieves only modest gains. Our method's consistent positive improvements across all imbalanced datasets highlight its robustness.

We attribute these improvements to our genetic algorithm's ability to generate completely balanced synthetic data with high intra-class diversity. 
While original datasets suffer from class imbalance, our crossover and mutation operations create diverse synthetic samples for each class, providing richer pattern variations that are particularly valuable for underrepresented classes. 
By providing abundant and diverse examples for each class, this balanced augmentation helps models develop better understanding of minority class patterns. This leads to more equitable performance across all classes, particularly benefiting applications where class imbalance is a persistent challenge.

\begin{table*}[ht]
\centering
\setlength{\tabcolsep}{4pt}
\resizebox{\textwidth}{!}{
\begin{tabular}{l|llllllll}
\toprule
\multirow{2}{*}{Method}
  & \multicolumn{1}{l}{AGnews}
  & \multicolumn{1}{l}{StackExchange}
  & \multicolumn{1}{l}{Chemprot}
  & \multicolumn{1}{l}{DDI}
  & \multicolumn{1}{l}{Semeval}
  & \multicolumn{1}{l}{Conll04}
  & \multicolumn{1}{l}{SciTLDR }
  & \multicolumn{1}{l}{MeQSum} \\
\cmidrule(lr){2-2}\cmidrule(lr){3-3}\cmidrule(lr){4-4}%
\cmidrule(lr){5-5}\cmidrule(lr){6-6}\cmidrule(lr){7-7}%
\cmidrule(lr){8-8}\cmidrule(lr){9-9}
 & Micro-F1& Micro-F1
    & Micro-F1  & Micro-F1  & Micro-F1  & Micro-F1  
    & Rouge-L & Rouge-L \\
\hline
Gold
                     & 91.9                     & 77.2                      & 82.1                      & 84.9                      & 92.4                      & 88.0                      & 29.6                      & 28.7                      \\
Gold + SimPrompt
                     &       91.5 {\color[HTML]{FF6347}(-0.4)}                & 78.6 {\color[HTML]{228B22}(+1.4)}                      & 83.8 {\color[HTML]{228B22}(+1.7)} & 83.8 {\color[HTML]{FF6347}(-1.1)} & 92.8 {\color[HTML]{228B22}(+0.4)} & 91.2 {\color[HTML]{228B22}(+3.2)} & 27.1 {\color[HTML]{FF6347}(-2.5)}                     & 24.4     {\color[HTML]{FF6347}(-4.3)}                \\
Gold + AttrPrompt
                    &92.0 {\color[HTML]{228B22}(+0.1)}                    & 79.9 {\color[HTML]{228B22}(+2.7)}                     & 83.0 {\color[HTML]{228B22}(+0.9)} & 84.6 {\color[HTML]{FF6347}(-0.3)} & 92.4 {\color[HTML]{228B22}(+0.0)} & 89.4 {\color[HTML]{228B22}(+1.4)} & 28.9  {\color[HTML]{FF6347}(-0.7)}                  & 28.2 {\color[HTML]{FF6347}(-0.5)}                     \\
Gold + Curated LLM
                     &  90.9 {\color[HTML]{FF6347}(-1.0)}                         & 78.0 {\color[HTML]{228B22}(+0.8)}                    & 83.9 {\color[HTML]{228B22}(+1.8)} & 85.4 {\color[HTML]{228B22}(+0.5)} & 92.2 {\color[HTML]{FF6347}(-0.2)} & 90.9 {\color[HTML]{228B22}(+2.9)} & -                     & -                    \\
Gold + Ours
                  & 92.3 {\color[HTML]{228B22}(+0.4)}                         & 80.0  {\color[HTML]{228B22}(+2.8)}                    & 84.4 {\color[HTML]{228B22}(+2.3)} & 85.9 {\color[HTML]{228B22}(+1.0)} & 92.9 {\color[HTML]{228B22}(+0.5)} & 92.1 {\color[HTML]{228B22}(+4.1)} & 30.1     {\color[HTML]{228B22}(+0.5)}                & 28.7       {\color[HTML]{228B22}(+0.0)}              \\
                \bottomrule
\end{tabular}%
}

\vspace{1em} 

\resizebox{\textwidth}{!}{%
\begin{tabular}{l|llllllll}
\toprule
\multirow{2}{*}{Method}
  & \multicolumn{1}{l}{AGnews}
  & \multicolumn{1}{l}{StackExchange}
  & \multicolumn{1}{l}{Chemprot}
  & \multicolumn{1}{l}{DDI}
  & \multicolumn{1}{l}{Semeval}
  & \multicolumn{1}{l}{Conll04}
  & \multicolumn{1}{l}{SciTLDR}
  & \multicolumn{1}{l}{MeQSum} \\
\cmidrule(lr){2-2}\cmidrule(lr){3-3}\cmidrule(lr){4-4}%
\cmidrule(lr){5-5}\cmidrule(lr){6-6}\cmidrule(lr){7-7}%
\cmidrule(lr){8-8}\cmidrule(lr){9-9}
  & Macro-F1 & Macro-F1 & Macro-F1 & Macro-F1 & Macro-F1 & Macro-F1 & Rouge-1 & Rouge-1 \\
\hline
Gold
                   & 91.9                      & 75.8                      & 60.7                      & 80.0                      & 92.0                      & 88.5                      & 33.4                     &   31.3                   \\
Gold + SimPrompt
                      & 91.4 {\color[HTML]{FF6347}(-0.5)}                      & 78.0 {\color[HTML]{228B22}(+2.2)}                     & 63.1 {\color[HTML]{228B22}(+2.4)} & 79.4 {\color[HTML]{FF6347}(-0.6)} & 92.5 {\color[HTML]{228B22}(+0.5)} & 91.6 {\color[HTML]{228B22}(+3.1)} & 31.5 {\color[HTML]{FF6347}(-1.9)}                      & 26.7 {\color[HTML]{FF6347}(-4.6)}                  \\
Gold + AttrPrompt
                      & 92.1 {\color[HTML]{228B22}(+0.2)}                      & 79.4 {\color[HTML]{228B22}(+3.6)}                     & 63.4 {\color[HTML]{228B22}(+2.7)} & 81.0 {\color[HTML]{228B22}(+1.0)} & 92.1 {\color[HTML]{228B22}(+0.1)} & 89.7 {\color[HTML]{228B22}(+1.2)} & 33.1 {\color[HTML]{FF6347}(-0.3)}                      & 31.0 {\color[HTML]{FF6347}(-0.3)}                    \\
Gold + Curated LLM
                      & 90.8 {\color[HTML]{FF6347}(-1.1)}                      & 77.7 {\color[HTML]{228B22}(+1.9)}                      & 64.0 {\color[HTML]{228B22}(+3.3)} & 82.3 {\color[HTML]{228B22}(+2.3)} & 92.0 {\color[HTML]{228B22}(+0.0)} & 91.1 {\color[HTML]{228B22}(+2.6)} & - & -                     \\
Gold + Ours
                      & 92.3 {\color[HTML]{228B22}(+0.4)}  & 79.5 {\color[HTML]{228B22}(+3.7)}  & 63.9 {\color[HTML]{228B22}(+3.2)} & 83.1 {\color[HTML]{228B22}(+3.1)} & 92.7 {\color[HTML]{228B22}(+0.7)} & 92.5 {\color[HTML]{228B22}(+4.0)} & 34.8 {\color[HTML]{228B22}(+1.4)}  & 31.7 {\color[HTML]{228B22}(+0.4)} \\
\bottomrule
\end{tabular}%
}
\caption{Gold–synthetic data fusion results with absolute percentage improvements. Top: Micro-F1 / Rouge-L. Bottom: Macro-F1 / Rouge-1.}
\label{tab:augmentation-combined}
\end{table*}

\begin{table*}[ht]
\setlength{\tabcolsep}{5pt}
\resizebox{1\textwidth}{!}{%
\begin{tabular}{l|cccccccc}
\toprule
\multirow{2}{*}{Method}
  
  & \multicolumn{1}{c}{AGnews}
  & \multicolumn{1}{c}{StackExchange}
  & \multicolumn{1}{c}{Chemprot}
  & \multicolumn{1}{c}{DDI}
  & \multicolumn{1}{c}{Semeval}
  & \multicolumn{1}{c}{Conll04}
  & \multicolumn{1}{c}{SciTLDR}
  & \multicolumn{1}{c}{MeQSum} \\
\cmidrule(lr){2-2}\cmidrule(lr){3-3}\cmidrule(lr){4-4}%
\cmidrule(lr){5-5}\cmidrule(lr){6-6}\cmidrule(lr){7-7}%
\cmidrule(lr){8-8}\cmidrule(lr){9-9}
& Micro-F1& Micro-F1
    & Micro-F1  & Micro-F1  & Micro-F1  & Micro-F1  
    & Rouge-L & Rouge-L \\
\hline
w/o Active Learning
   & 82.4 & 50.0 & 71.7 & 67.7 & 68.8 & 69.3 & 22.7 & 19.0 \\
w/o Mutation
   & 81.9 & 49.4 & 74.0 & 67.3 & 70.8 & 51.4 & 26.0 & 24.1 \\
Word as Gene
   & 71.5 & 40.5 & 52.7 & 54.0 & 54.4 & 56.6 & 22.5 & 23.2 \\
Ours
   & \textbf{84.0} & \textbf{51.2} & \textbf{77.2} & \textbf{68.7} & \textbf{72.3} & \textbf{72.8} & \textbf{27.5} & \textbf{27.4} \\
\bottomrule
\end{tabular}%
}
\caption{Ablation study of the individual modules.}
\label{tab:ablation}
\end{table*}

\subsection{Ablation Study}

To assess contributions of the individual modules within our Genetic Prompt framework, we conducted an ablation study by: 1) replacing the active learning-based parent selection with a random selection mechanism while allowing parent reuse (``w/o Active Learning''), 2) entirely removing the mutation operation (``w/o Mutation''), and 3) replacing the semantic genes with a traditional word-level gene (``Word as Gene'').
For the "Word as Gene" configuration, LLMs are not required to understand the semantic features of parent sequences. Instead, they utilize linguistic features to perform crossover and mutation operations.
We perform the experiments on the Llama3.1-70B.

As shown in Table~\ref{tab:ablation}, our complete method consistently outperforms all ablated versions across all datasets.
Without the mutation operation, the LLM is constrained in its ability to introduce sufficient perturbations during offspring generation, thereby limiting the exploration of the generative search space and reducing offspring diversity. 
Similarly, the absence of the active learning mechanism makes the parent selection process to rely on random sampling, which may repeatedly select highly similar instances. 
Consequently, as these similar offspring are incorporated into the population, the likelihood of selecting highly similar parents in subsequent generations increases, thereby reinforcing the cycle of performance decline.
Finally, the ``Word as Gene'' module focusing on word-level crossover and mutation significantly restricts the generative process and reduces downstream NLP models. 
The finding verifies our assumption and analysis in Section~\ref{sec:method} that our semantic-level approach is more effective than word-level augmentation for high-quality synthetic data; and our semantic-level genetic algorithm may enhance data diversity and lead to downstream improvements.

\section{Effects Analysis of Generator Size and Data Scale}
\label{sec:Empirical Analysis}

\begin{figure*}[htbp]
    \centering
    \begin{subfigure}[b]{\textwidth}
        \centering
        \begin{minipage}{0.32\textwidth}
            \centering
            \includegraphics[width=\textwidth]{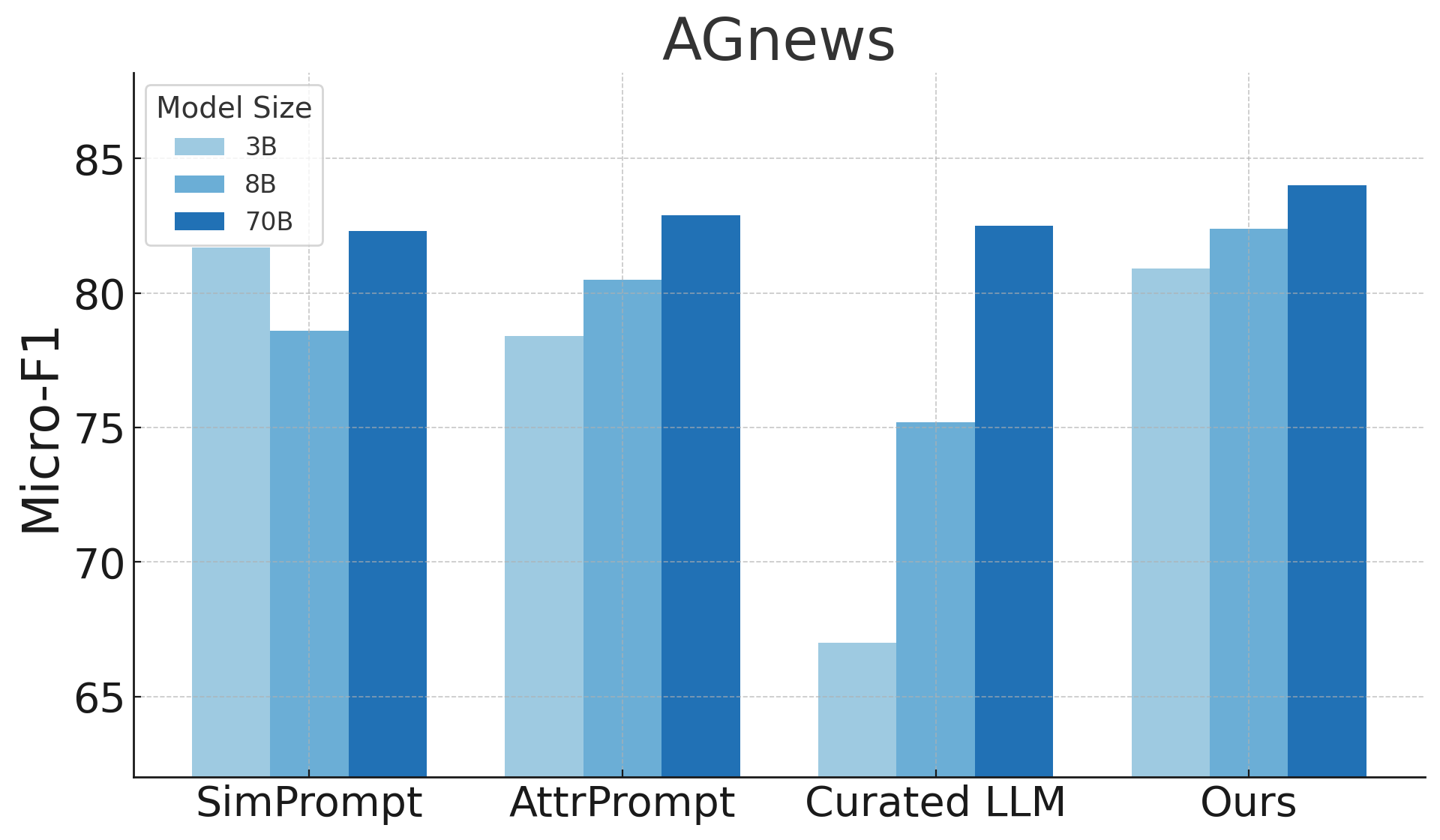}
        \end{minipage}
        \hfill
        \begin{minipage}{0.32\textwidth}
            \centering
            \includegraphics[width=\textwidth]{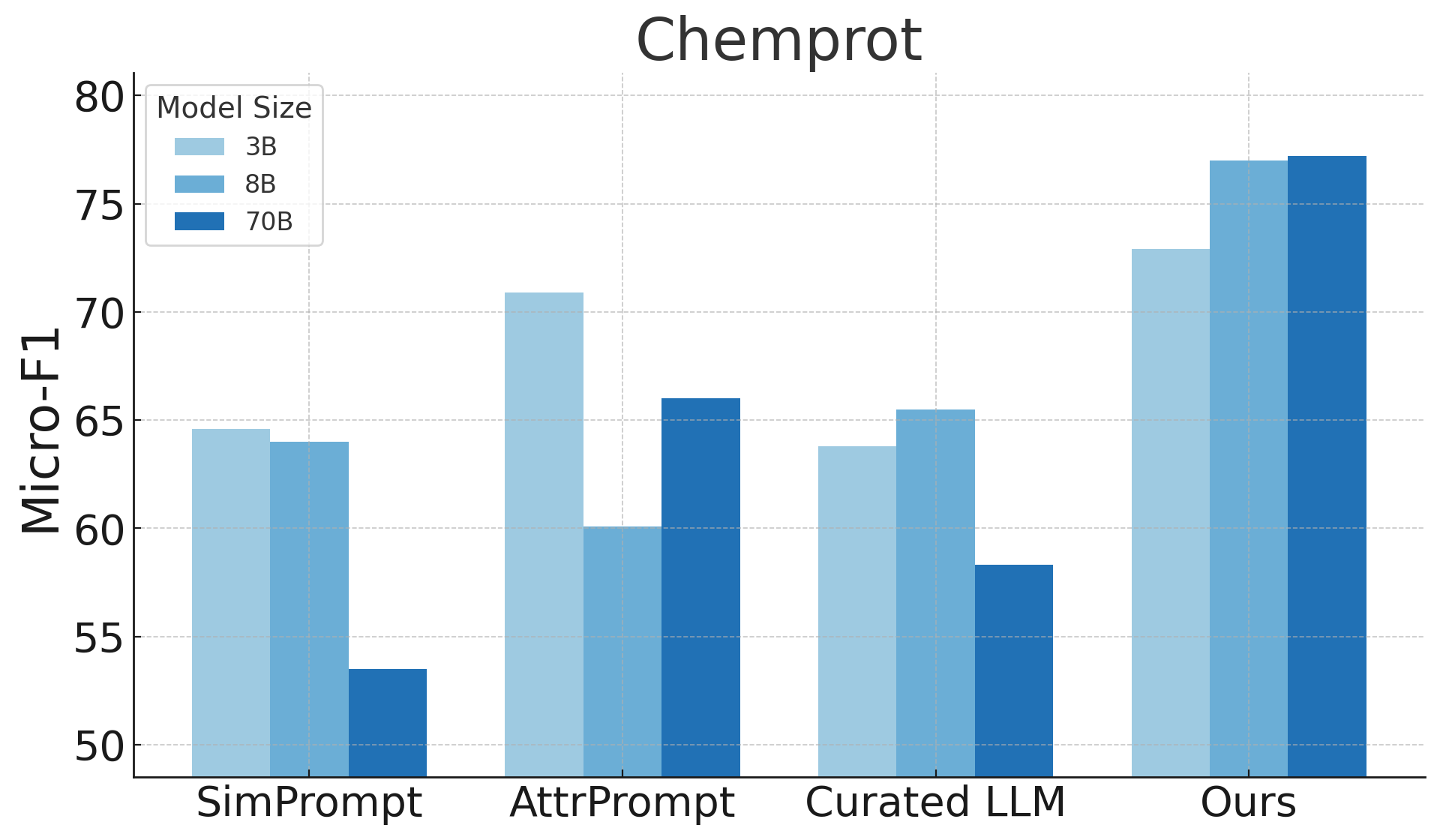}
        \end{minipage}
        \hfill
        \begin{minipage}{0.32\textwidth}
            \centering
            \includegraphics[width=\textwidth]{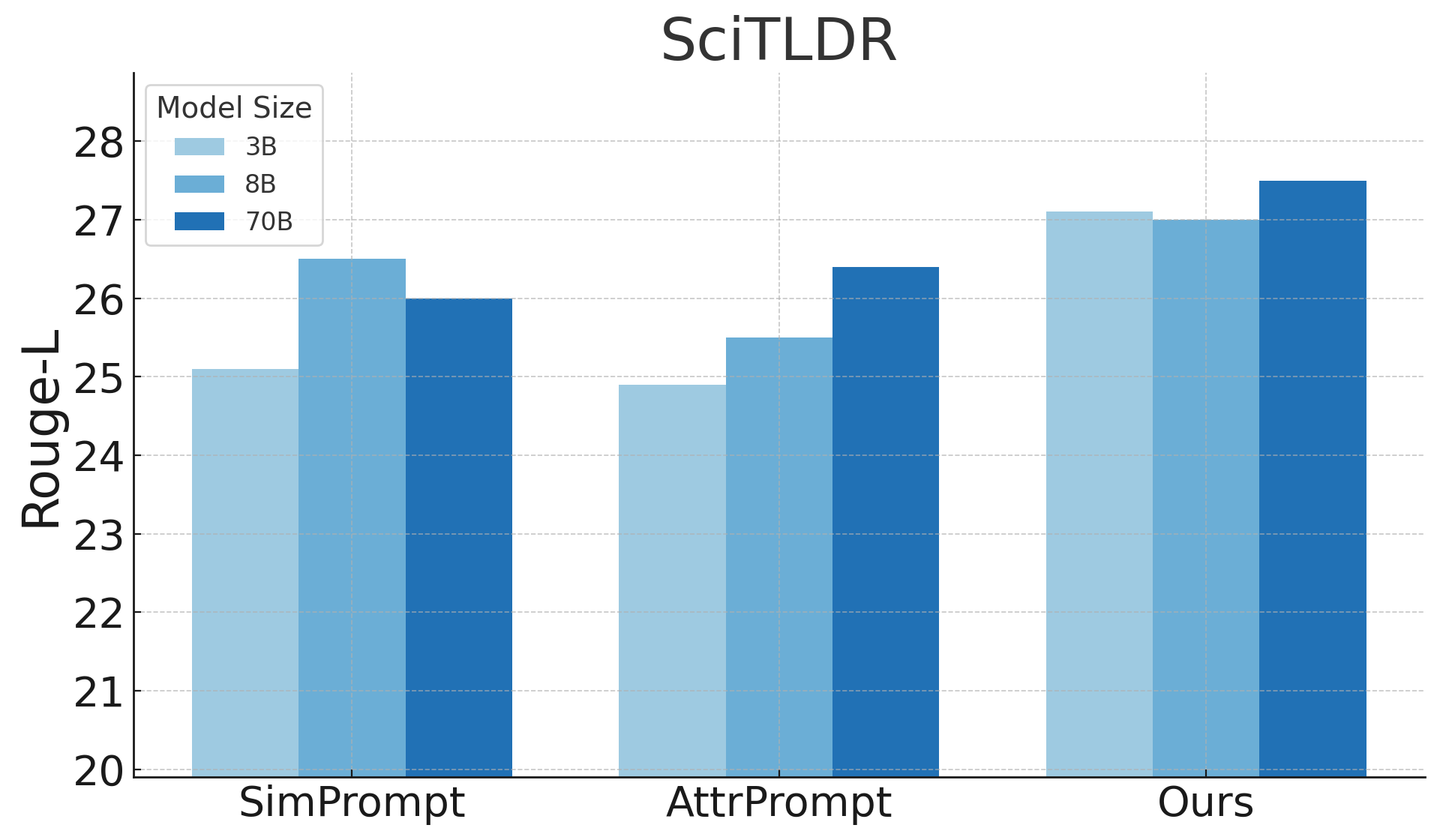}
        \end{minipage}
        \caption{Effects of generator size (Llama3.2-3B, Llama3.1-8B, Llama3.1-70B) across three tasks: text classification (AGNews), relation extraction (ChemProt), and text summarization (SciTLDR).}
        \label{fig:generator_size}
    \end{subfigure}
    
    \vspace{1em}
    
    \begin{subfigure}[b]{\textwidth}
        \centering
        \begin{minipage}{0.32\textwidth}
            \centering
            \includegraphics[width=\textwidth]{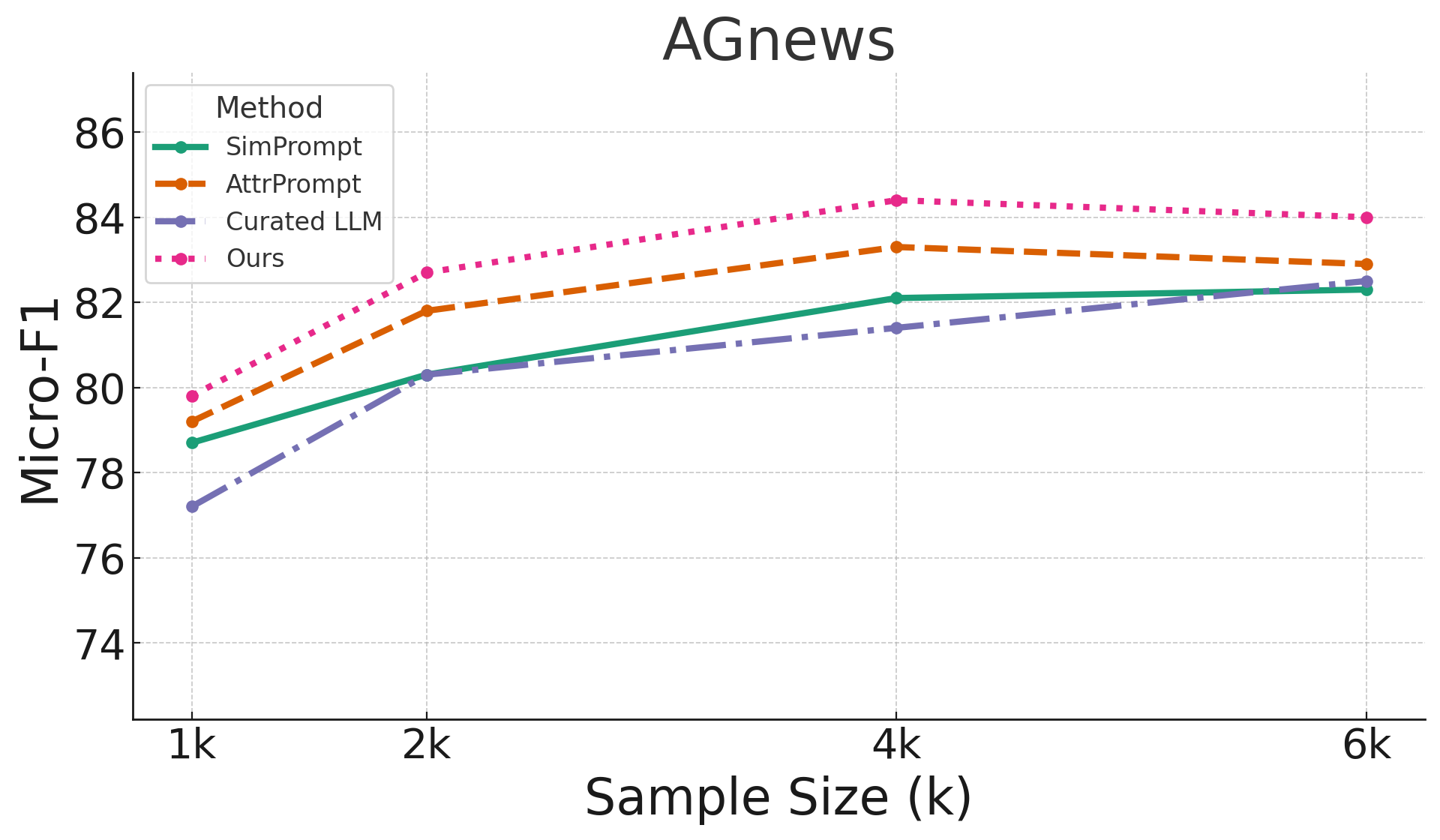}
        \end{minipage}
        \hfill
        \begin{minipage}{0.32\textwidth}
            \centering
            \includegraphics[width=\textwidth]{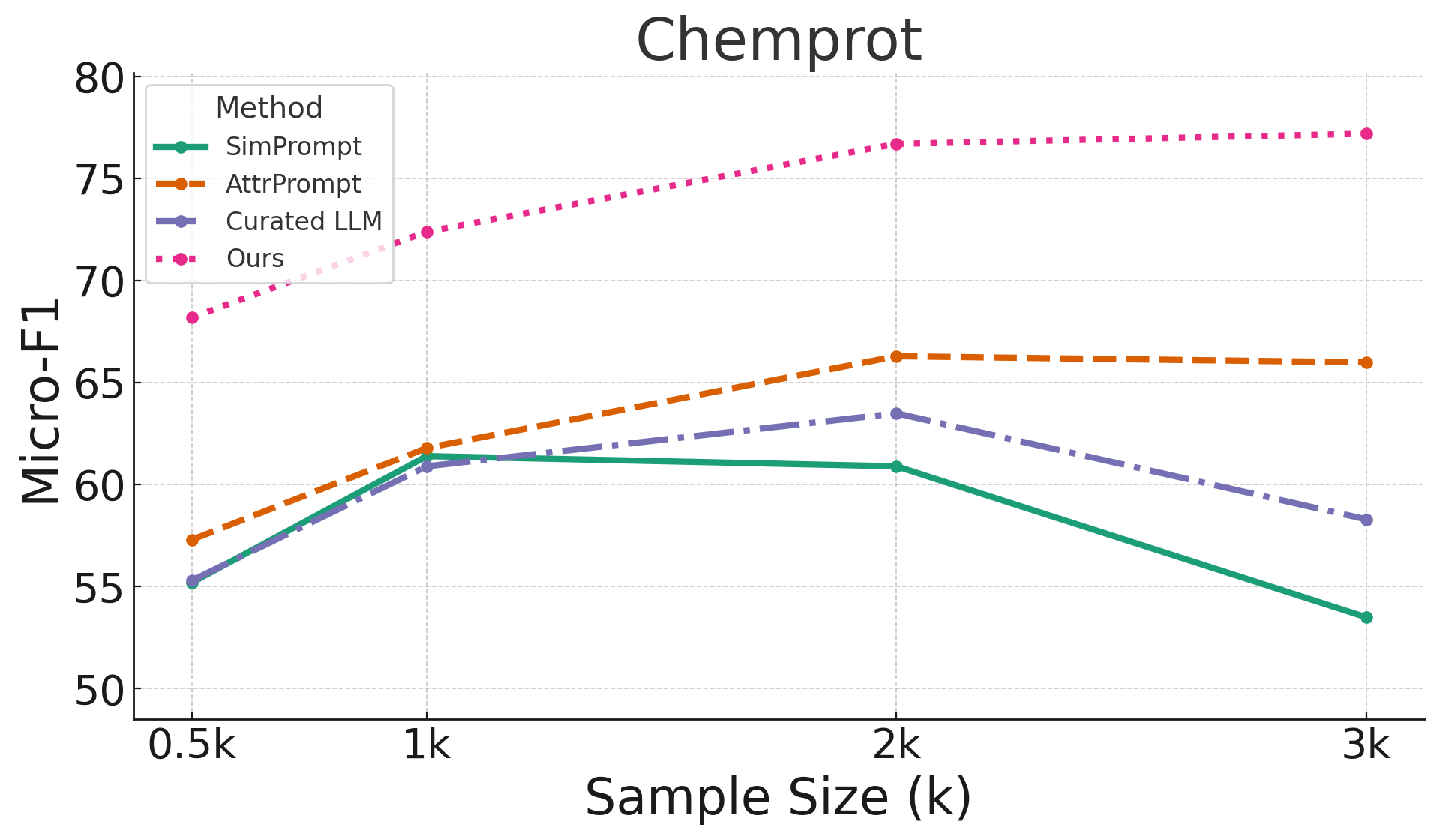}
        \end{minipage}
        \hfill
        \begin{minipage}{0.32\textwidth}
            \centering
            \includegraphics[width=\textwidth]{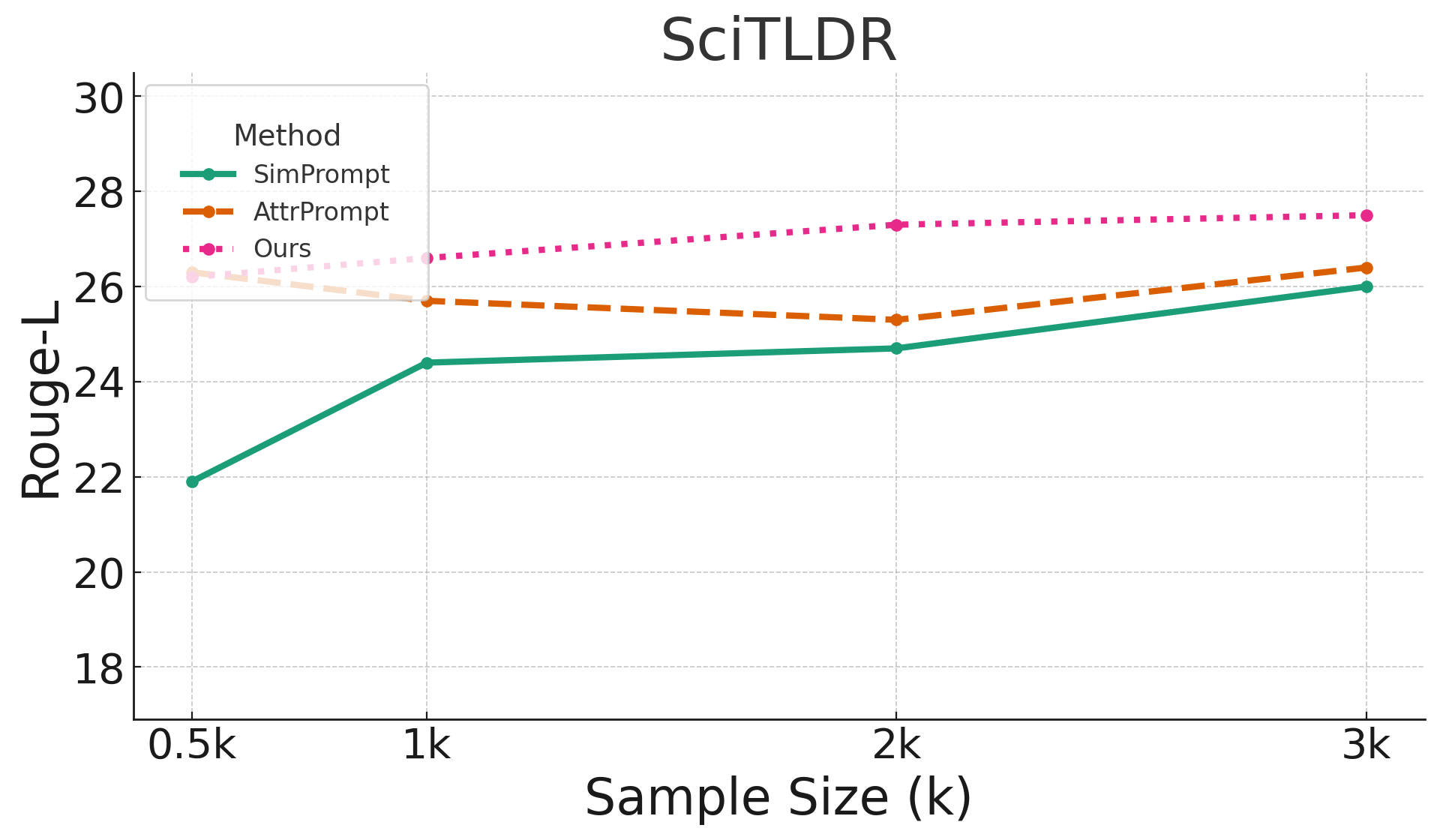}
        \end{minipage}
        \caption{Effects of synthetic data scale across three tasks: text classification (AGNews), relation extraction (ChemProt), and text summarization (SciTLDR).}
        \label{fig:data_scale}
    \end{subfigure}
    
    \caption{Effects analysis of generator size and synthetic data scale across diverse NLP tasks.}
    \label{fig:Empirical Analysis}
\end{figure*}

As data varies in Table~\ref{tab:data} in scale and sizes, thus a concrete question to be examined is: how may data size and scale impact on synthetic data generation and their downstream applications? 
To answer the question, we conducted experiments on three models that share the same architecture, training methods, and even the same training data distribution: Llama3.2-3B, Llama3.1-8B, and Llama3.1-70B.
We report performance variations across data scale and size in Figure~\ref{fig:Empirical Analysis} with other tasks and baseline results in Figure~\ref{fig:appendix-3-8-70} and Figure~\ref{fig:appendix-1k-2k-3k} under Appendix.

\paragraph{Effects of Data Generator Size.}

As shown in Figure~\ref{fig:generator_size}, different datasets exhibit varying sensitivity to generator model size. On the ChemProt data (relation extraction), performance of our Genetic Prompt positively correlates with the size of the Llama models, while SimPrompt and Curated LLM exhibit the opposite trend, and AttrPrompt shows a decrease in performance using the middle-size Llama model. On the AGNews data (text classification), methods show large performance gaps on smaller models (3B, 8B) but converge to similar performance levels on the largest model (70B), suggesting that sufficient model capacity can compensate for methodological differences in this task. 
On the SciTLDR data (text summarization), moderate sensitivity is observed with some performance fluctuations across model sizes.
These observations reveal that the impact of generator size is highly task-dependent. 
Larger generators do not universally lead to improved performance on downstream tasks, while our method demonstrates consistent robustness across models of various sizes, with notable advantages on tasks that are more sensitive to model scale such as relation extraction.

\paragraph{Effects of Synthetic Data Scale.}
Our Genetic Prompt demonstrates consistent improvement as synthetic data size increases across all three tasks in Figure~\ref{fig:data_scale}. 
In contrast, other methods show different scaling patterns: on AGNews, all methods steadily improve with increasing data scale; on ChemProt, SimPrompt, AttrPrompt, and Curated LLM plateau or decline after reaching peak performance around 1-2k samples; on SciTLDR, most methods show relatively stable performance with modest improvements.

We also observe two key patterns across different methods and data.
First, although different methods achieve varying peak performance levels on each dataset, they tend to reach their optimal performance at similar data scales per task. 
Second, different datasets exhibit significantly different optimal synthetic data scales despite having comparable task complexity - for instance, while AGNews (4 classes) and ChemProt (5 classes) have similar numbers of categories, AGNews reaches peak performance around 4k samples while ChemProt peaks at approximately 2k samples.

The superior scalability of our method, particularly evident on ChemProt where other methods suffer from performance degradation at larger scales, suggests better data diversity and reduced overfitting in our generated samples. This is consistent with the results in Table~\ref{tab:diversity}.

\section{Related Work}
\label{sec:related work}
\subsection{Genetic Algorithm with LLMs}

LLMs possess deep domain knowledge and text processing capabilities, while genetic algorithms excel at global search and iterative optimization. 
To harness these complementary strengths, the main collaboration paradigms involve either using LLMs to enhance genetic algorithms or employing genetic algorithms to refine LLM outputs~\cite{wu2024evolutionary,Justyna2018Genetic}. 
These approaches have been broadly applied in code generation~\cite{lehman2022evolution}, summary generation, and neural network architecture search~\cite{chen2023evoprompting}. 
For instance, LLMs can serve as evolution strategies~\cite{robert2024large} or evolutionary operators~\cite{guo2024connecting,yang2024large} in black-box prompt optimization. 
Furthermore, if sentences or paragraphs are viewed as gene sequences~\cite{liu2024autodan}, one can employ a hierarchical genetic algorithm on a set of prompt prefixes to identify the optimal prompt prefix for a given task.

However, these works share common limitations: they rely on syntactic crossover and mutation operations at the word or sentence level, converging toward a single optimal output~\cite{liu2024autodan}. 
Therefore, this focus on finding one best candidate limits the ability to maintain both quality and diversity when generating large volumes of text.
To address these shortcomings, our Genetic Prompt framework shifts the notion of ``genes'' from syntactic units (words or sentences) to higher-level semantic attributes, such as writing style and sentence structure. 
In doing so, we harness the comprehension and reasoning capabilities of LLMs to perform crossover and mutation, enabling us to simultaneously achieve high quality and diversity at scale. 
Moreover, instead of relying on a conventional fitness-based parent selection process, which can be difficult to evaluate at the individual sample level, we integrate an active learning approach~\cite{zhang2023llmaaa,Ren2021survey-AL}. 
In each iteration, we pair previously unused, diverse parents to expand offspring search space, thereby enhancing both the quality and variety of the generated text.

\subsection{Synthetic Data Generation with LLMs}

The emergence of LLMs has established them as key drivers of synthetic data generation across diverse domains~\cite{long2024llms}. 
These applications span multiple areas, including text classification tasks~\cite{han2024chain,xu2024knowledge,liu2025examining}, tabular data synthesis~\cite{zhang2025features,yang2025doubling}, retrieval query generation~\cite{lu2025multiconir}, benchmark construction~\cite{zhang2025CII-Bench,wu2024unigen}, and LLM post-training~\cite{chen2025consistentchat,zhou2025reso,lan2025contextual,wang2025code,wang2025vl}.

However, simply relying on LLM-generated data neither guarantees quality and diversity~\cite{yue2023large} nor prevents the introduction of biases inherent to these models. 
To overcome these limitations, various enhancement strategies have been proposed, including multi-step generation~\cite{cui2024adainstruct}, in-context learning~\cite{li2024selfprompting,xu2024knowledge}, label augmentation~\cite{xiao2023freeal}, and conditional prompting~\cite{yu2021attribute,russo2020control,logeswaran2018content}. 
For instance, some approaches extract and predefine text attributes~\cite{yue2023large} or integrate external knowledge~\cite{xu2024knowledge,rao2025apt}, leveraging this information as constraints to guide data generation. 
While these conditional prompting methods effectively raise the baseline level of text diversity, they also introduce drawbacks. 
In particular, imposing numerous constraints can hinder an LLM’s reasoning and comprehension, limiting its ability to learn from context and accurately reflect real-world patterns and distributions.

To address these issues, Genetic Prompt abandons explicit conditional constraints. 
Instead, it treats the semantic attributes of text as genes in a genetic algorithm, thereby better leveraging the LLM's inherent knowledge and comprehension abilities to generate high-quality data from a limited set of real examples. 
Moreover, by incorporating crossover and mutation operations, Genetic Prompt introduces sufficient perturbations to enhance the diversity and scalability of the synthetic data, effectively reducing the risk of overfitting in downstream models.

\section{Conclusion}
\label{sec:conclusion}

In this work, we introduced the Genetic Prompt Framework, a novel approach combining LLMs and genetic algorithms for synthetic data generation. 
Experiments on 8 datasets from multiple domains and NLP tasks demonstrate the effectiveness of Genetic Prompt, as it consistently outperformed three state-of-the-art baselines across multiple settings. 
Our ablation and effect analysis have summarized multiple insights for future studies on generating synthetic data or applying synthetic data over downstream tasks: 1) semantic-level generation of synthetic data can capture more data characteristics and thus promote synthetic data quality, 2) combining synthetic and real-world data generally boost model performance but rely on LLM options and prompting methods, and 3) synthetic data play a more critical role in downstream tasks with smaller and domain-specific data.
Our future work will further expand and verify the potentials to apply our approach to broader domains, modalities, and applications, such as synthetic generation for health tabular data and training small language models.

\section*{Limitations}
\label{sec:limitations}

In this study, we propose Genetic Prompt for LLMs to generate synthetic data.
Despite the strong performance and good data quality achieved, several limitations should be acknowledged:
First, our experiments were conducted exclusively on English corpora. Other languages may possess distinct linguistic features and characteristics that differ from English, potentially affecting the quality of synthetic data generation. The effectiveness of our approach across diverse languages remains to be validated.
Second, our approach was specifically designed for textual data synthesis. Extending this work to other modalities such as tabular, or to multimodal data synthesis scenarios would likely require substantial adaptations to the methodology.

\section*{Acknowledgment}
The authors thank anonymous reviewers for their insightful feedback. 
This project was supported by the National Science Foundation (NSF) IIS-2245920 and CNS-2318210.
We thank for the computing resources provided by the iTiger GPU cluster\footnote{\url{https://itiger-cluster.github.io/}}~\cite{sharif2025ITIGER} supported by the NSF MRI program.
We would also thank for additional funding from the ITS and College of Arts and Sciences at the University of Memphis to partially support daily HPC operations.

\bibliography{custom}

\appendix

\section{Additional Results}

\label{sec:appendix-additional-results}
\subsection{Supplementary Experimental Results}

Due to space limitations, we present the remaining performance metrics of the three experiments \textbf{Extrinsic Evaluation on NLP Tasks} and \textbf{Ablation Study} in Table~\ref{tab:main-results-macro-extended} and Table~\ref{tab:ablation-macro} respectively.

\begin{table*}[ht]
\centering
\setlength{\tabcolsep}{2pt}
\resizebox{\textwidth}{!}{%
\begin{tabular}{ll|cccccccc}
\toprule
\multirow{2}{*}{Method} & \multirow{2}{*}{Model}
  
  & \multicolumn{1}{c}{AGnews} 
  & \multicolumn{1}{c}{StackExchange} 
  & \multicolumn{1}{c}{Chemprot} 
  & \multicolumn{1}{c}{DDI} 
  & \multicolumn{1}{c}{Semeval} 
  & \multicolumn{1}{c}{Conll04} 
  & \multicolumn{1}{c}{SciTLDR} 
  & \multicolumn{1}{c}{MeQSum} \\
\cmidrule(lr){3-3}\cmidrule(lr){4-4}\cmidrule(lr){5-5}%
\cmidrule(lr){6-6}\cmidrule(lr){7-7}\cmidrule(lr){8-8}%
\cmidrule(lr){9-9}\cmidrule(lr){10-10}
 & & Macro-F1 & Macro-F1 & Macro-F1 & Macro-F1 & Macro-F1 & Macro-F1 & Rouge-1 & Rouge-1 \\
\hline
\multirow{4}{*}{SimPrompt}
 & Phi4           & 78.3$_{0.5}$ & 45.9$_{0.5}$ & 53.7$_{1.9}$  & 44.7$_{1.5}$  & 55.3$_{2.0}$  & 63.5$_{2.0}$  & 29.5$_{0.1}$ & 25.2$_{0.3}$ \\
 & Llama3.1-70b     & 82.0$_{1.5}$ & 50.2$_{0.3}$ & 42.6$_{1.7}$  & 56.5$_{1.6}$  & 53.4$_{1.2}$  & 53.0$_{5.7}$  & 29.8$_{0.1}$ & 24.4$_{0.3}$ \\
 & GPT-3.5-Turbo  & 70.8$_{0.5}$ & 39.8$_{0.2}$ & 53.4$_{1.3}$  & 63.3$_{1.4}$  & 62.4$_{0.5}$  & 77.2$_{1.9}$  & 30.2$_{0.3}$ & 25.3$_{0.2}$ \\
 & GPT-4o         & 81.7$_{1.1}$ & 50.7$_{1.2}$ & 58.1$_{3.0}$  & 56.3$_{2.1}$  & 66.5$_{0.9}$  & 71.1$_{2.2}$  & 30.0$_{0.2}$ & 24.8$_{0.1}$ \\
\hline
\multirow{4}{*}{AttrPrompt}
 & Phi4           & 78.9$_{2.0}$ & 49.1$_{0.2}$ & 52.1$_{0.4}$  & \textbf{68.3}$_{0.8}$  & 61.2$_{1.3}$  & 59.3$_{2.0}$  & 30.1$_{0.1}$ & 27.0$_{0.0}$ \\
 & Llama3.1-70b     & 82.2$_{0.4}$ & 47.4$_{0.8}$ & 51.8$_{0.8}$  & 62.3$_{0.9}$  & 66.0$_{0.7}$  & 64.9$_{1.9}$  & 30.5$_{0.0}$ & 26.8$_{0.2}$ \\
 & GPT-3.5-Turbo  & 79.7$_{0.7}$ & 47.0$_{0.7}$ & 56.1$_{1.9}$  & 63.7$_{2.5}$  & 70.0$_{0.5}$  & 67.8$_{2.8}$  & 30.7$_{0.2}$ & 27.1$_{0.2}$ \\
 & GPT-4o         & 80.7$_{1.5}$ & 48.3$_{0.3}$ & 57.7$_{2.1}$  & 60.7$_{2.3}$  & 70.0$_{1.1}$  & 73.1$_{3.9}$  & 30.2$_{0.1}$ & 27.9$_{0.2}$ \\
\hline
\multirow{4}{*}{Curated LLM}
 & Phi4          & 78.9$_{0.4}$ & 43.8$_{1.0}$ & 57.1$_{2.2}$  & 52.2$_{1.4}$  & 53.1$_{2.1}$  & 63.8$_{3.0}$  & - & - \\
 & Llama3.1-70b    & 82.0$_{1.1}$ & 48.5$_{0.4}$ & 44.8$_{3.3}$  & 65.4$_{2.0}$  & 52.2$_{0.8}$  & 49.0$_{1.4}$  & - & - \\
 & GPT-3.5-Turbo  & 77.2$_{0.3}$ & 42.1$_{0.4}$ & 55.6$_{1.3}$  & 65.2$_{1.2}$  & 62.5$_{1.0}$  & 73.1$_{1.5}$  & - & - \\
 & GPT-4o         & 81.6$_{1.6}$ & 49.2$_{0.6}$ & 49.6$_{2.9}$  & 55.7$_{2.9}$  & 67.9$_{0.2}$  & 74.7$_{2.4}$  & - & - \\
\hline
\multirow{4}{*}{Ours}
 & Phi4           & 80.1$_{1.3}$ & \textbf{54.6}$_{0.2}$ & 56.4$_{1.9}$  & 63.1$_{0.7}$  & 65.2$_{0.5}$  & \textbf{72.9}\textbf{$_{2.9}$} & 31.3$_{0.2}$ & \textbf{29.2}$_{0.1}$ \\
 & Llama3.1-70b     & \textbf{83.4}$_{1.0}$ & 49.9$_{1.0}$ & \textbf{58.9}\textbf{$_{1.2}$} & 67.6\textbf{$_{2.0}$} & \textbf{71.3}\textbf{$_{1.5}$} & 72.5$_{1.9}$  & \textbf{31.6}$_{0.2}$ & 29.2$_{0.2}$ \\
 & GPT-3.5-Turbo  & 82.5$_{0.5}$ & 49.3$_{0.3}$ & 62.5$_{1.0}$  & \textbf{69.2$_{2.0}$} & 71.0$_{1.0}$  & 66.7$_{3.5}$  & 30.4$_{0.1}$ & 27.3$_{0.1}$ \\
 & GPT-4o         & \textbf{86.5}$_{0.7}$ & 48.0$_{0.3}$ & \textbf{65.4$_{1.9}$} & 62.6$_{3.0}$  & \textbf{78.1$_{0.5}$} & \textbf{85.8$_{1.2}$} & \textbf{32.1}$_{0.2}$ & 29.0$_{0.2}$ \\
\bottomrule
\end{tabular}}
\caption{Additional experimental results of the Genetic Prompt framework and baselines. We \textbf{bold} the best overall performance and the best performance among open-source LLMs. Standard deviations are calculated using three random seeds.}
\label{tab:main-results-macro-extended}
\end{table*}

\begin{table*}[ht]

\resizebox{\textwidth}{!}{%
\begin{tabular}{l|cccccccc}
\toprule
\multirow{2}{*}{Method}
  
  & \multicolumn{1}{c}{AGnews}
  & \multicolumn{1}{c}{StackExchange}
  & \multicolumn{1}{c}{Chemprot}
  & \multicolumn{1}{c}{DDI}
  & \multicolumn{1}{c}{Semeval}
  & \multicolumn{1}{c}{Conll04}
  & \multicolumn{1}{c}{SciTLDR}
  & \multicolumn{1}{c}{MeQSum} \\
\cmidrule(lr){2-2}\cmidrule(lr){3-3}\cmidrule(lr){4-4}%
\cmidrule(lr){5-5}\cmidrule(lr){6-6}\cmidrule(lr){7-7}%
\cmidrule(lr){8-8}\cmidrule(lr){9-9}
  & Macro-F1 & Macro-F1 & Macro-F1 & Macro-F1 & Macro-F1 & Macro-F1 & Rouge-1 & Rouge-1 \\
\hline
w/o Active Learning
   & 82.3 & 49.3 & 57.3 & 66.2 & 67.6 & 69.2 & 26.2 & 21.3 \\
w/o Mutation
   & 81.8 & 49.0 & 53.8 & 58.6 & 69.8 & 40.8 & 30.1 & 26.4 \\
Word as Gene
  & 70.6 & 38.3 & 46.7 & 47.7 & 52.2 & 50.5 & 27.4 & 24.9 \\
Ours
   & \textbf{83.4} & \textbf{49.9} & \textbf{58.9} & \textbf{67.6} & \textbf{71.3} & \textbf{72.5} & \textbf{31.6} & \textbf{29.2} \\
\bottomrule
\end{tabular}%
}
\caption{Additional ablation study results.}
\label{tab:ablation-macro}
\end{table*}

\subsection{The Effect of Data Generator Size}

\begin{figure}[htbp]
  \centering
  
  \begin{subfigure}[b]{0.49\textwidth}
    \centering
    \includegraphics[width=\linewidth]{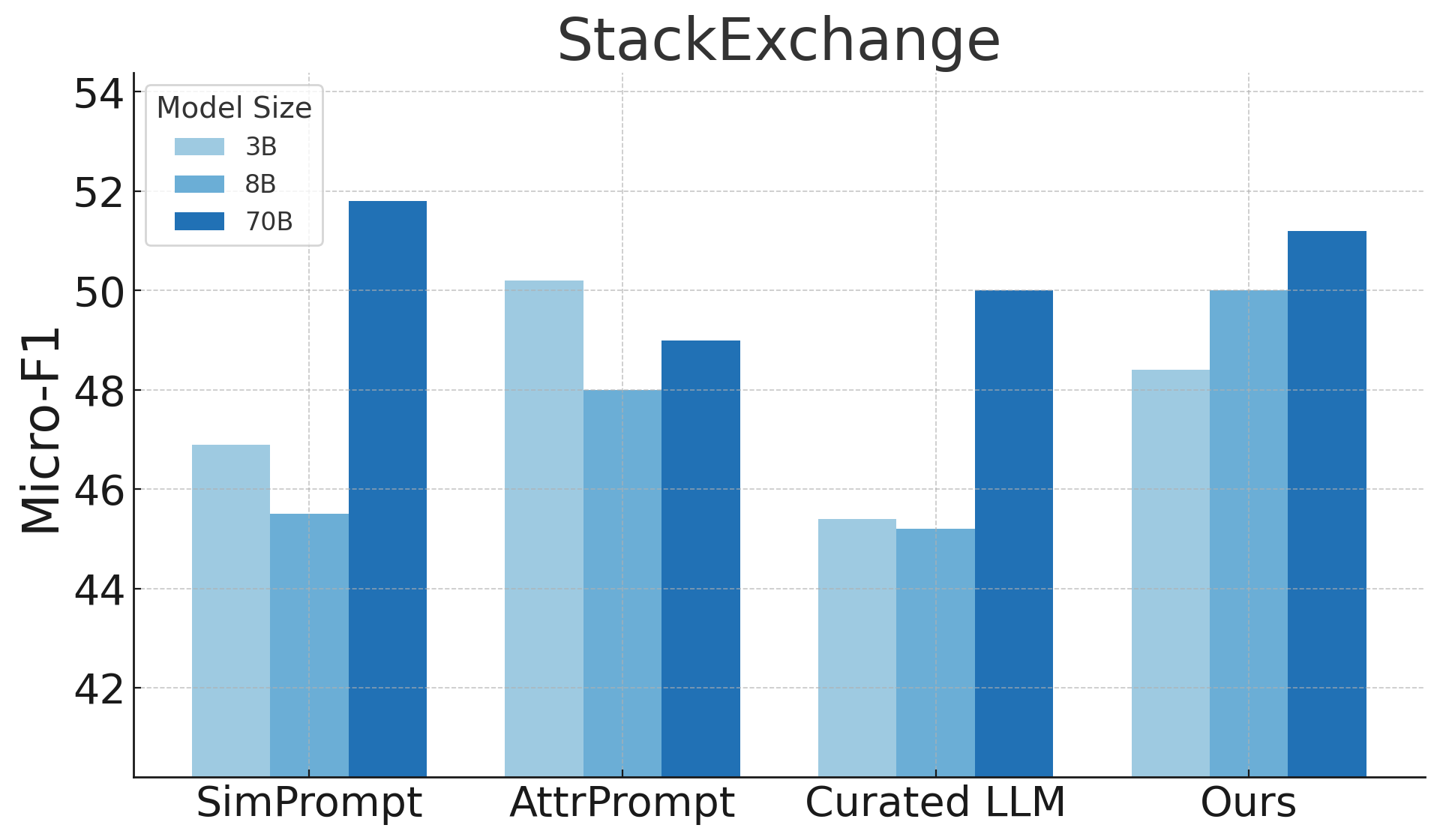}
  \end{subfigure}
  
  \vspace{0.5em}
  
  \begin{subfigure}[b]{0.49\textwidth}
    \centering
    \includegraphics[width=\linewidth]{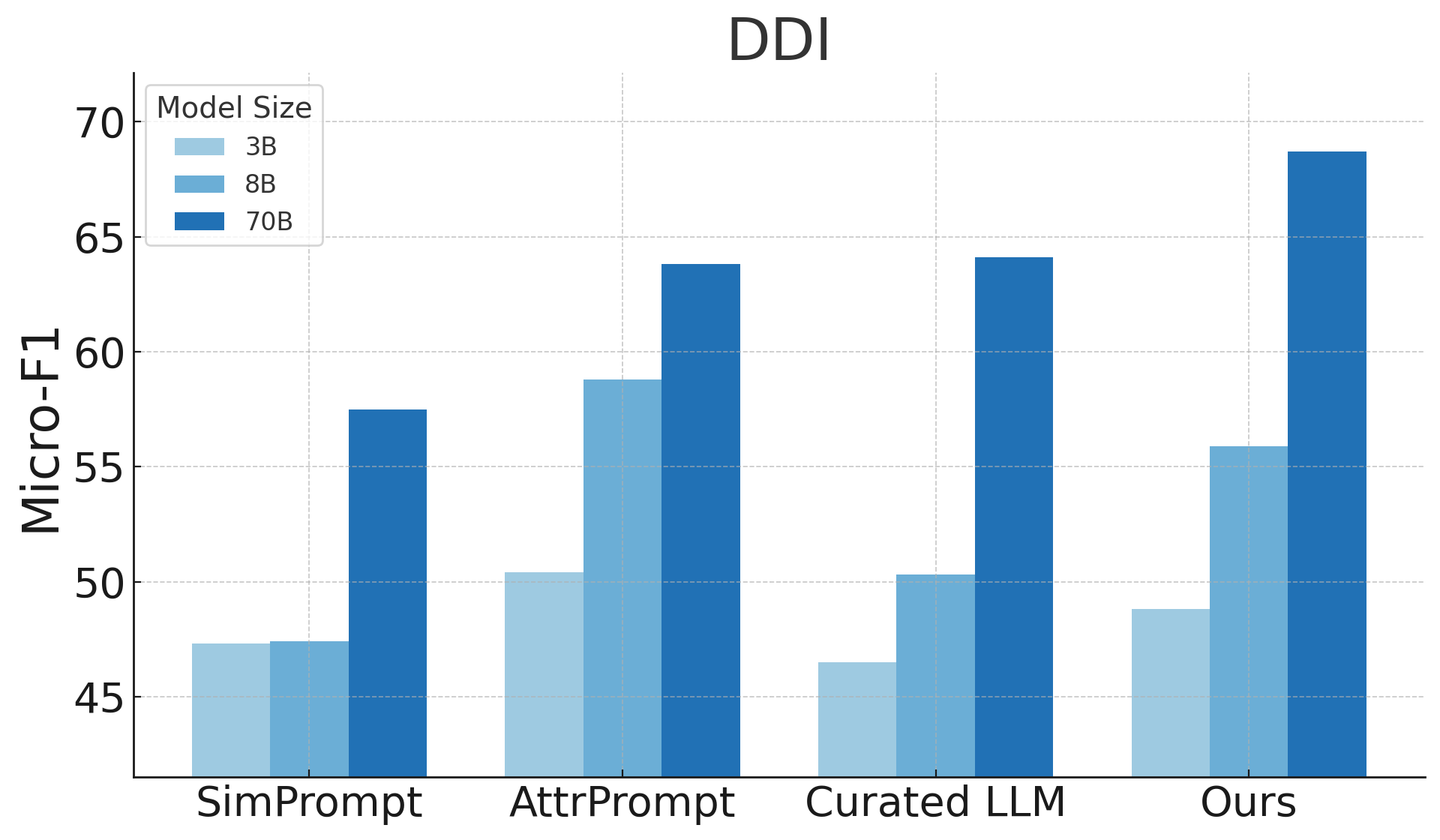}
  \end{subfigure}
  
  \vspace{0.5em}
  
  \begin{subfigure}[b]{0.49\textwidth}
    \centering
    \includegraphics[width=\linewidth]{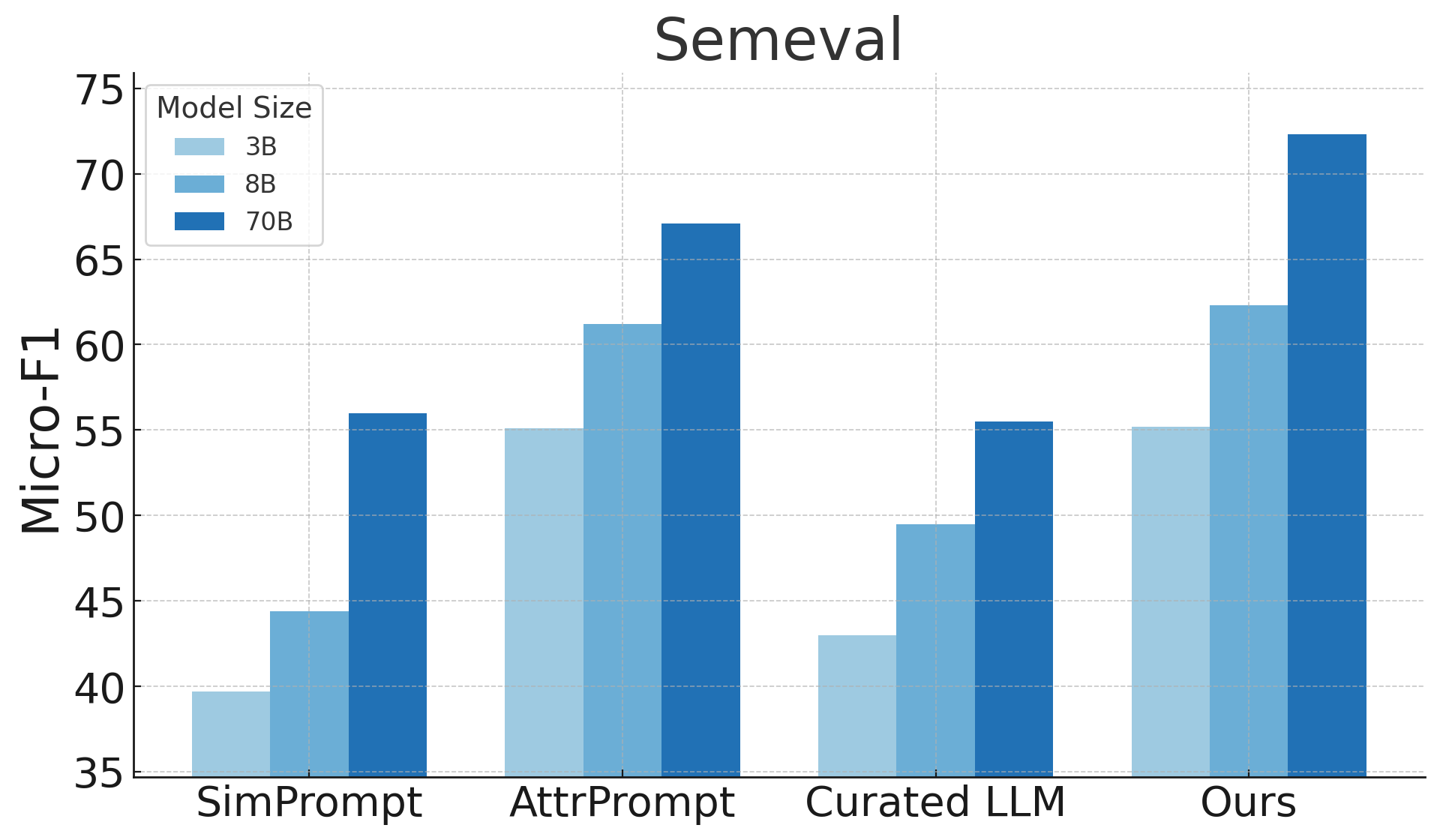}
  \end{subfigure}
  
  \vspace{0.5em}
  
  \begin{subfigure}[b]{0.49\textwidth}
    \centering
    \includegraphics[width=\linewidth]{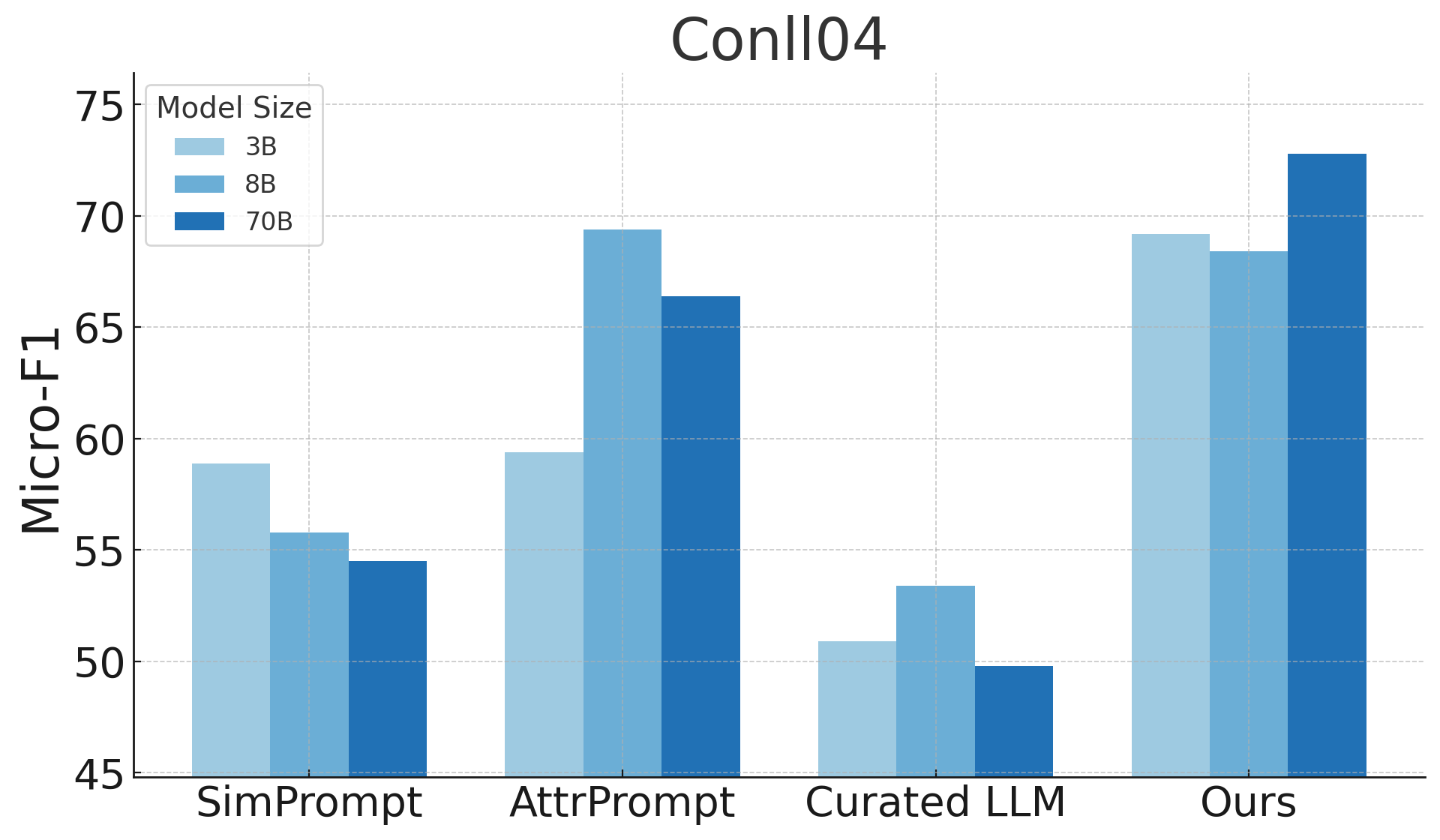}
  \end{subfigure}
  
  \vspace{0.5em}
  
  \begin{subfigure}[b]{0.49\textwidth}
    \centering
    \includegraphics[width=\linewidth]{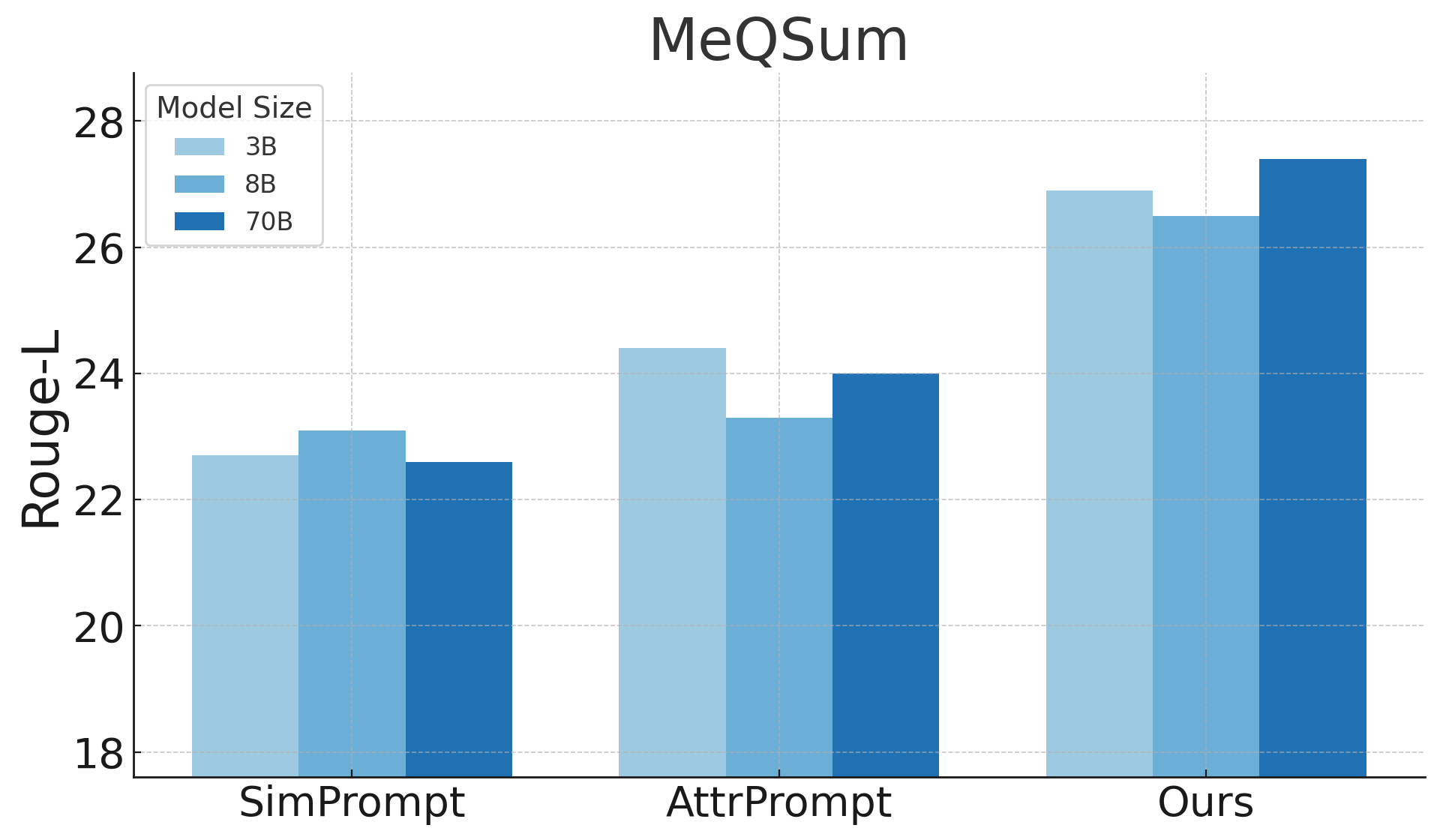}
  \end{subfigure}
  
  \caption{Empirical analysis on generator size}
  \label{fig:appendix-3-8-70}
\end{figure}

For same space reason, we report additional results of empirical analysis on generator size in Table~\ref{tab:llama-scale} and Figure~\ref{fig:appendix-3-8-70}.

\begin{sidewaystable}[ht]
\centering
\setlength{\tabcolsep}{3pt}
\resizebox{\textwidth}{!}{%
\begin{tabular}{l| *{8}{ccc}}
\toprule
\multirow{2}{*}{\textbf{Method}} & \multicolumn{3}{c}{\textbf{AGNews}} & \multicolumn{3}{c}{\textbf{StackExchange}} & \multicolumn{3}{c}{\textbf{Chemprot}} & \multicolumn{3}{c}{\textbf{DDI}} & \multicolumn{3}{c}{\textbf{Semeval}} & \multicolumn{3}{c}{\textbf{Conll}} & \multicolumn{3}{c}{\textbf{SciTLDR}} & \multicolumn{3}{c}{\textbf{MeQSum}} \\
\cmidrule(lr){2-4} \cmidrule(lr){5-7} \cmidrule(lr){8-10} \cmidrule(lr){11-13} \cmidrule(lr){14-16} \cmidrule(lr){17-19} \cmidrule(lr){20-22} \cmidrule(lr){23-25}
 & 3B & 8B & 70B & 3B & 8B & 70B & 3B & 8B & 70B & 3B & 8B & 70B & 3B & 8B & 70B & 3B & 8B & 70B & 3B & 8B & 70B & 3B & 8B & 70B \\
\midrule
SimPrompt   & \textbf{81.7} & 78.6 & 82.3 & 46.9 & 45.5 & \textbf{51.8}   & 64.6 & 64.0 & 53.5 & 47.3 & 47.4 & 57.5 & 39.7 & 44.4 & 56.0 & 58.9 & 55.8 & 54.5 & 25.1 & 26.5 & 26.0 & 22.7 & 23.1 & 22.6 \\
AttrPrompt   & 78.4 & 80.5 & 82.9 & \textbf{50.2} & 48.0 & 49.0 & 70.9 & 60.1 & 66.0 & \textbf{50.4} & \textbf{58.8} & 63.8 & 55.1 & 61.2 & 67.1 & 59.4 & \textbf{69.4} & 66.4 & 24.9 & 25.5 & 26.4 & 24.4 & 23.3 & 24.0 \\
Curated LLM & 67.0 & 75.2 & 82.5 & 45.4 & 45.2 & 50.0 & 63.8 & 65.5 & 58.3 & 46.5 & 50.3 & 64.1 & 43.0 & 49.5 & 55.5 & 50.9 & 53.4 & 49.8 & - & - & - & - & - & - \\
Ours   & 80.9 & \textbf{82.4} &\textbf{ 84.0} & 48.4 & \textbf{50.0} & 51.2     & \textbf{72.9} & \textbf{77.0} & \textbf{77.2} & 48.8 & 55.9 & \textbf{68.7} & \textbf{55.2} & \textbf{62.3} & \textbf{72.3} & \textbf{69.2} & 68.4 & \textbf{72.8} & \textbf{27.1} & \textbf{27.0} & \textbf{27.5} & \textbf{26.9} & \textbf{26.5} & \textbf{27.4} \\
\bottomrule
\end{tabular}%
}
\caption{Additional results of empirical analysis on generator size. We report Rouge-1 and Macro-F1 for generation and classification tasks separately.}
\label{tab:llama-scale}
\end{sidewaystable}

\subsection{The Effect of Synthetic Data Scale}
\begin{figure}[htbp]
  \centering
  
  \begin{subfigure}[b]{0.49\textwidth}
    \centering
    \includegraphics[width=\linewidth]{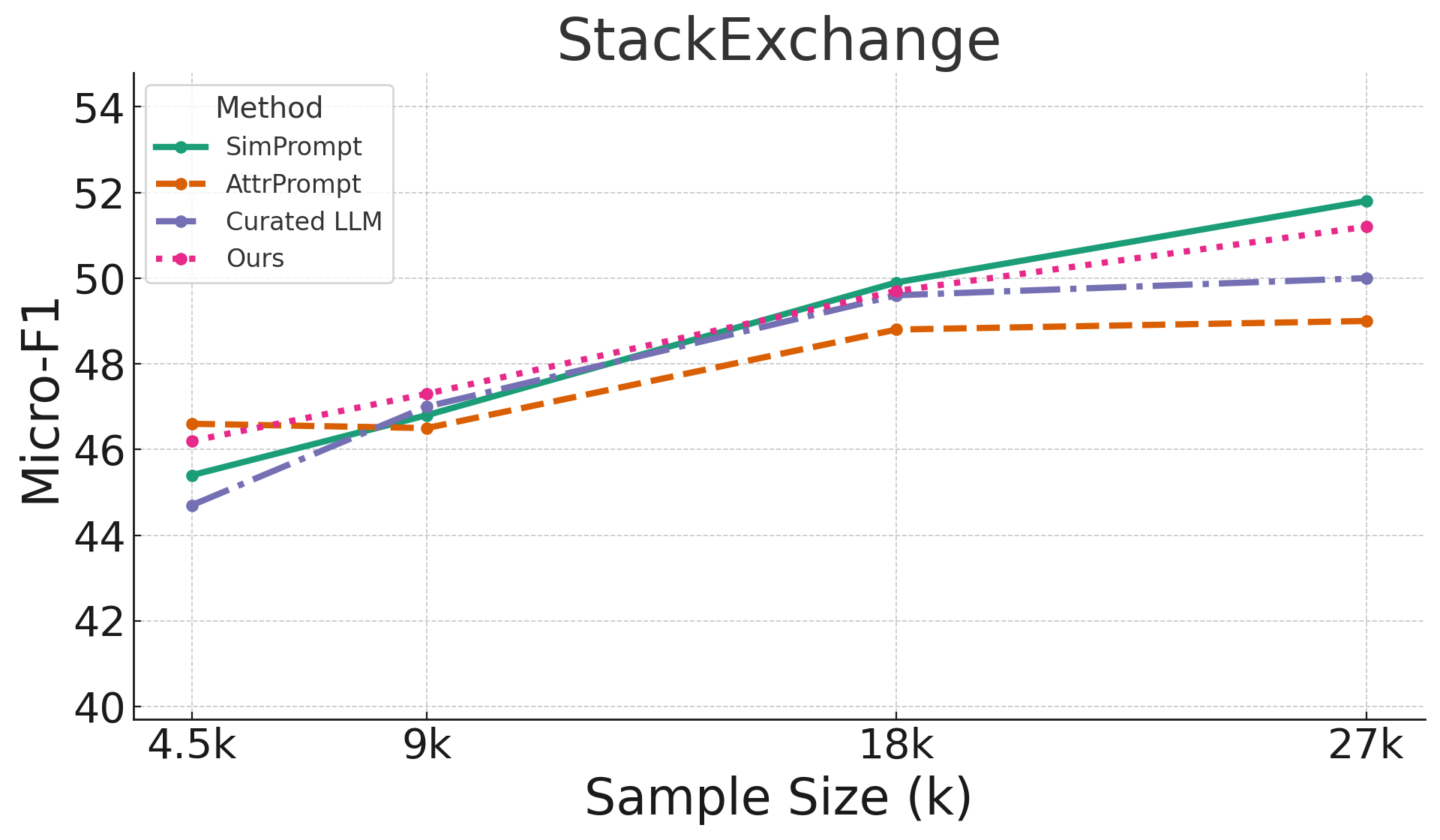}
  \end{subfigure}
  
  \vspace{0.5em}
  
  \begin{subfigure}[b]{0.49\textwidth}
    \centering
    \includegraphics[width=\linewidth]{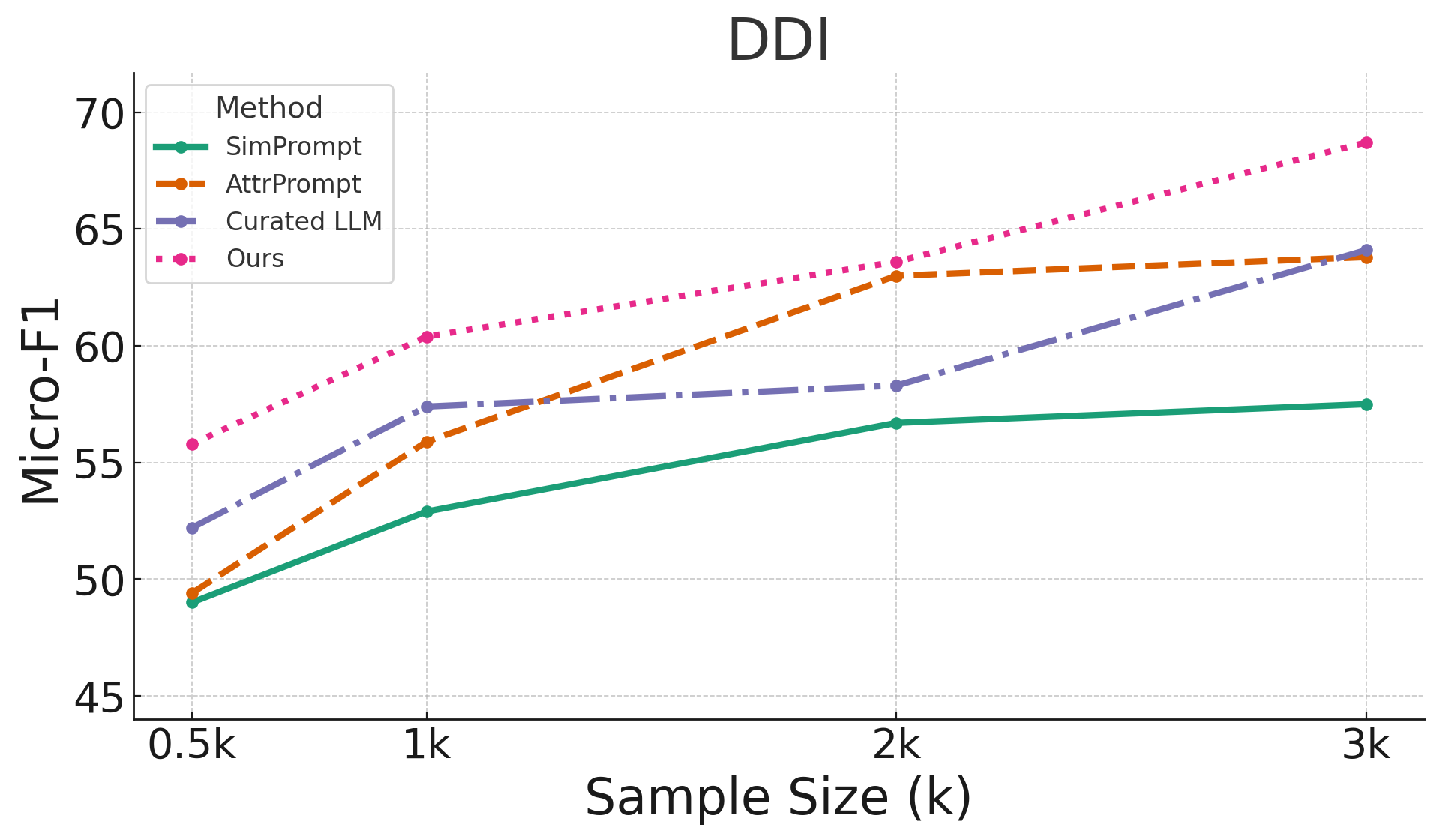}
  \end{subfigure}
  
  \vspace{0.5em}
  
  \begin{subfigure}[b]{0.49\textwidth}
    \centering
    \includegraphics[width=\linewidth]{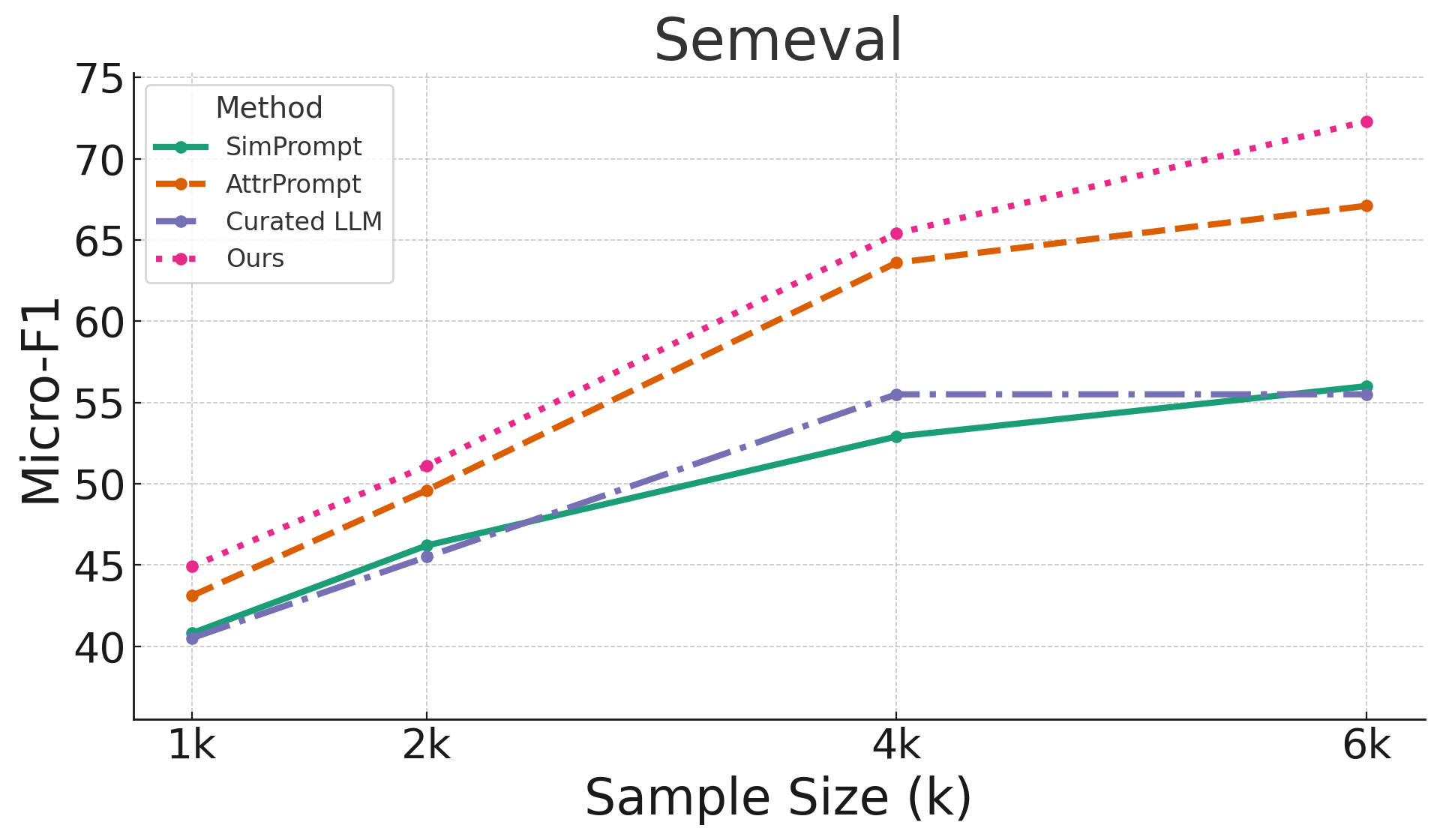}
  \end{subfigure}
  
  \vspace{0.5em}
  
  \begin{subfigure}[b]{0.49\textwidth}
    \centering
    \includegraphics[width=\linewidth]{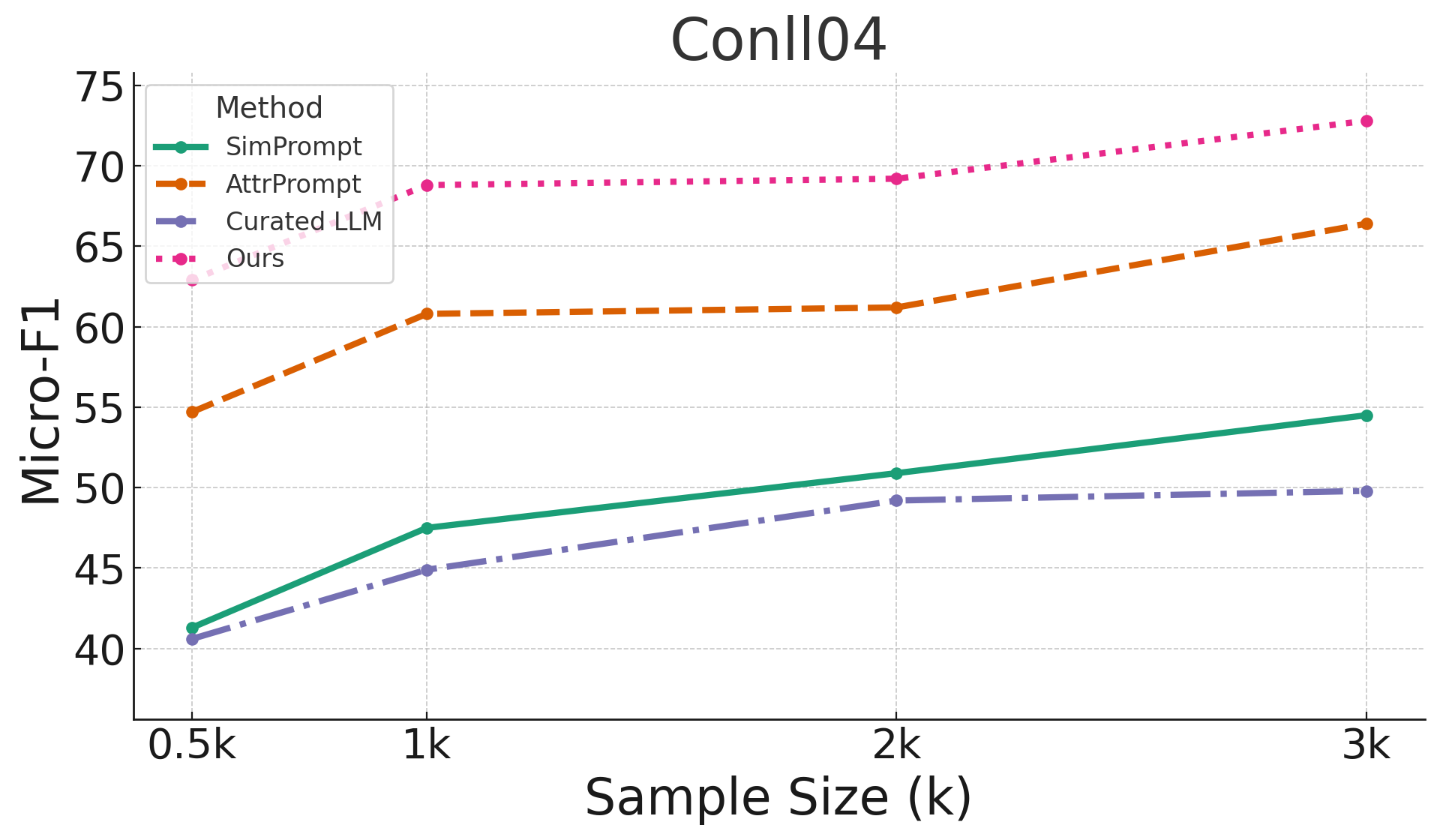}
  \end{subfigure}
  
  \vspace{0.5em}
  
  \begin{subfigure}[b]{0.49\textwidth}
    \centering
    \includegraphics[width=\linewidth]{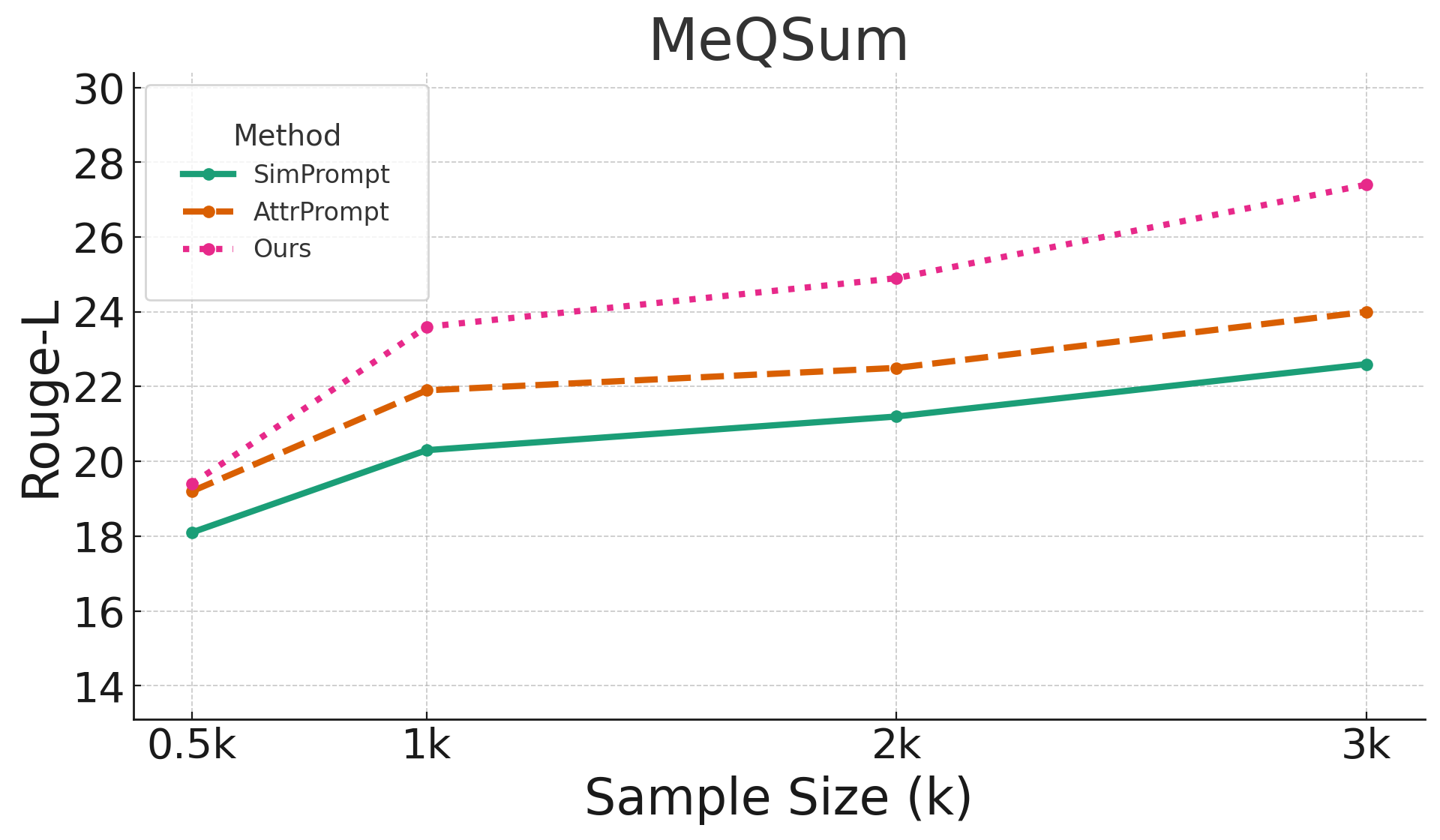}
  \end{subfigure}
  
  \caption{Empirical analysis on sample size.}
  \label{fig:appendix-1k-2k-3k}
\end{figure}

we report additional results of empirical analysis on data scale in Table~\ref{tab:6k} and Figure~\ref{fig:appendix-1k-2k-3k}.

\begin{sidewaystable}[ht]
\centering
\setlength{\tabcolsep}{2pt}
\resizebox{\textwidth}{!}{%
\begin{tabular}{l|cccccccccccccccccccccccccccccccc}
\toprule
\multirow{2}{*}{\textbf{Method}} & \multicolumn{4}{c}{\textbf{AGNews}} & \multicolumn{4}{c}{\textbf{StackExchange}} & \multicolumn{4}{c}{\textbf{Chemprot}} & \multicolumn{4}{c}{\textbf{DDI}} & \multicolumn{4}{c}{\textbf{Semeval}} & \multicolumn{4}{c}{\textbf{Conll04}} & \multicolumn{4}{c}{\textbf{SciTLDR}} & \multicolumn{4}{c}{\textbf{MeQSum}} \\
\cmidrule(lr){2-5} \cmidrule(lr){6-9} \cmidrule(lr){10-13} \cmidrule(lr){14-17} \cmidrule(lr){18-21} \cmidrule(lr){22-25} \cmidrule(lr){26-29} \cmidrule(lr){30-33}
                & 1k & 2k  & 4k  & 6k  & 4.5k & 9k  & 18k  & 27k & 0.5k & 1k  & 2k  & 3k  & 0.5k & 1k  & 2k  & 3k  & 1k & 2k  & 4k  & 6k  & 0.5k & 1k  & 2k  & 3k & 0.5k & 1k  & 2k  & 3k & 0.5k & 1k  & 2k  & 3k \\
\midrule
SimPrompt   & 78.7 & 80.3 & 82.1 & 82.3 & 45.4 & 46.8 & \textbf{49.9} & \textbf{51.8} & 55.2 & 61.4 & 60.9 & 53.5 & 49.0 & 52.9 & 56.7 & 57.5 & 40.8 & 46.2 & 52.9 & 56.0 & 41.3 & 47.5 & 50.9 & 54.5 & 21.9 & 24.4 & 24.7 & 26.0 & 18.1 & 20.3 & 21.2 & 22.6 \\
AttrPrompt   & 79.2 & 81.8 & 83.3 & 82.9 & \textbf{46.6 }& 46.5 & 48.8 & 49.0 & 57.3 & 61.8 & 66.3 & 66.0 & 49.4 & 55.9 & 63.0 & 63.8 & 43.1 & 49.6 & 63.6 & 67.1 & 54.7 & 60.8 & 61.2 & 66.4 & \textbf{26.3} & 25.7 & 25.3 & 26.4 & 19.2 & 21.9 & 22.5 & 24.0 \\
Curated LLM & 77.2 & 80.3 & 81.4 & 82.5 & 44.7 & 47.0 & 49.6 & 50.0 & 55.3 & 60.9 & 63.5 & 58.3 & 52.2 & 57.4 & 58.3 & 64.1 & 40.5 & 45.5 & 55.5 & 55.5 & 40.6 & 44.9 & 49.2 & 49.8 & - & - & - & - & - & - & - & - \\
Ours   & \textbf{79.8} & \textbf{82.7} & \textbf{84.4 }& \textbf{84.0} & 46.2 & \textbf{47.3} & 49.7 & 51.2 & \textbf{68.2} & \textbf{72.4} & \textbf{76.7} & \textbf{77.2} & \textbf{55.8} & \textbf{60.4} &\textbf{ 63.6} & \textbf{68.7} & \textbf{44.9} & \textbf{51.1} & \textbf{65.4} & \textbf{72.3} & \textbf{62.9} & \textbf{68.8} & \textbf{69.2} & \textbf{72.8} & 26.2 & \textbf{26.6} & \textbf{27.3} & \textbf{27.5} & \textbf{19.4 }& \textbf{23.6} & \textbf{24.9} & \textbf{27.4} \\
\bottomrule
\end{tabular}%
}
\caption{Additional results of empirical analysis on data scale. We report Rouge-1 and Macro-F1 for generation and classification tasks separately.}
\label{tab:6k}
\end{sidewaystable}

\clearpage

\clearpage

\begin{algorithm*}
\caption{Genetic Algorithm Implemented by LLM for Synthetic Data Generation}
\begin{algorithmic}[1]
    \State \textbf{Input}: Label set $L$, textual genes $G$, target data size $N$, diversity coefficient $\alpha$.

    \State Initialize population $P=\{p_0,p_1,\ldots,p_{50}\}$ from gold dataset for each label $l$.
    \State Build a sample pool $\{P,E\}$ for each label $l$ for the initial population. 
    
    \Comment $E$ are semantic representations of population $P$.
    \State Calculate the diversity score $APS=\{a_0,a_1,\ldots,a_{l}\}$ of each class in gold dataset.
    
    \Comment{APS: \textbf{A}verage \textbf{P}air wise sample \textbf{S}imilarity.}
    \For{$l \in L$}
    \While{population $<N$ }
        \State Get semantic distances of pairs in $\{P,E\}$: 
        $\forall e_i, e_j \in E, \quad d_{ij} = \text{dist}(e_i, e_j) $
        \State Select the most distant pair from \textbf{unused} samples as parents: $(p_i^*, p_j^*) = \argmax_{e_i, e_j} d_{ij}$

        \State Randomly partition textual genes $G$ into three groups: $G = G_1 \cup G_2 \cup G_3$.
        \State Perform LLM-based crossover on $G_1$ and $G_2$ using parents $(p_i^*, p_j^*)$ and perform mutation on $G_3$ : $p_{new} = \text{LLM-GA}(p_i^*, p_j^*, G_1, G_2, G_3)$
        \State Add the new sample to pool: $\{P,E\} := \{P,E\} \cup \{p_{new}, e_{new}\}$

    \EndWhile
    \State Remove the initial population from the sample pool.
    \EndFor
\State \textbf{Output}: Synthetic data 
\end{algorithmic}
\label{alg:LLM-GA}
\end{algorithm*}

\begin{table*}[htbp]
    \centering
    \resizebox{\textwidth}{!}{
    \begin{tabular}{ll}
    \hline
    \textbf{Dataset} & \textbf{Genes} \\
    \hline

    AGnews & length, location, style, subtopics\\
    StackExchange & length, style, depth, scenario\\
    Chemprot    & length, voice, sentence structure, interaction verb, modifier, negation, entity proximity \\
    DDI         & length, voice, polarity, interaction verb, modifier, drug mentions, entity proximity \\
    SemEval2010 & length, voice, sentence structure, readability, entity distance, domain \\
    CoNLL04     & length, voice, sentence structure, readability, entity distance, domain \\
    SciTLDR & length, depth, style, subtopics\\
    MeQSum &length, depth, style, subtopics\\
    \hline
    \end{tabular}}
    \caption{Selected Textual Genes (Attributes) for Our Genetic Prompt and baseline AttrPrompt.}
    \label{tab:textual_genes}
\end{table*}

\clearpage

\section{Implementation Details}
\label{sec:appendix-implementation details}

\subsection{Datasets}
\label{appendix:data}

\paragraph{AGNews~\cite{agnews}} contains 120,000 categorized news articles from more than 2000 news sources. 
\paragraph{StackExchange~\cite{geigle2021tweac}} contains structured technical content from knowledge-sharing platforms, including questions, answers, and user interactions across specialized domains.

\paragraph{Chemprot~\cite{chemprot} }is a biomedical dataset derived from PubMed abstracts, focusing on chemical-protein interaction relation extraction, containing sentences annotated with chemical and protein entities and their various fine-grained interaction types. We follow previous work~\cite{peng2019transfer}, use the same 5 classes: Substrate, Upregulator, Downregulator, Agonist and Antagonist.

\paragraph{DDI (Drug-Drug Interaction)~\cite{DDI} } is from the semeval-2013-task9 challenge, we filter all sentences that contain drug-drug interactions and follow ~\cite{peng2019transfer} to evaluate the performance of the four types of DDI.

\paragraph{Semeval2010-task8~\cite{SEMEVAL2010}} is a multi-Way classification task of Semantic Relations, which contains nine directed relations and the \texttt{Other} relation. 
To balance the relation types in the selected four datasets, We merge the 9 directed relation pairs into nine undirected symmetric relations and remove the \texttt{Other} relation.

\paragraph{Conll04~\cite{conll04}} is a relation extraction dataset consisting of news articles with annotations for four entity types and five relation types (Lives-In, Located-In, OrgBased-In, Work-For, and Kill). 
We  utilize the standard training and test sets for evaluating the performance of five relation types. 

\paragraph{SciTLDR~\cite{cachola2020tldr}} is a publicly available dataset of over 23,000 scientific papers paired with concise, one-sentence “TL;DR” summaries. 
In this work, we use their \textbf{Abstract-only} setting.
\paragraph{MeQSum~\cite{ben2019summarization-MeQSum}} a publicly available dataset of consumer health questions from the MedHelp forum paired with expert-written abstractive summaries.

\subsection{Additional Technical Details}
During generation, for responses that do not adhere to the instructions, such as \textit{`I am just a large language model.'} or \textit{`I cannot do that for you.' }, are filtered out.
The hyperparameters can be found in Table~\ref{tab:hyper}.
\begin{table}[ht]
  \centering

  \resizebox{\columnwidth}{!}{%
    \begin{tabular}{lllll}
      \toprule
      \textbf{Dataset}      & \textbf{Model}      & \textbf{Epoch} & \textbf{Learning Rate} & \textbf{Size} \\
      \midrule
      AGNews                & RoBERTa-base        & 5              & 1e-5     &6k              \\
      StackExchange         & RoBERTa-base        & 5              & 1e-5     &27k              \\
      Chemprot              & RoBERTa-base        & 3              & 1e-5      &3k             \\
      DDI                   & RoBERTa-base        & 3              & 1e-5      &3k             \\
      Semeval               & RoBERTa-base        & 3              & 1e-5       &6k            \\
      Conll04               & RoBERTa-base        & 3              & 1e-5        &3k           \\
      SciTLDR               & T5-large            & 10           & 3e-5        &3k           \\
      MeQSum                & T5-large            & 10           & 3e-5         &3k          \\
      \bottomrule
    \end{tabular}%
  }\caption{Hyperparameters of main experiments.}
  \label{tab:hyper}
\end{table}

\subsection{Hardware and Software}

We conducted all experiments on a machine equipped with 8x H100 GPUs, 2x EPYC Genoa 9334 CPUs, and 768GB of RAM. 
The system runs on Linux kernel 5.14.
Besides, for our proposed Genetic Prompt framework, we utilize the FAISS library~\cite{Johnson2021billion} for accelerate calculations. 
For the Sentence Transformer library~\cite{reimers2019sentence}  used in our experiments, we utilize the \texttt{stsb-roberta-large} checkpoint as the encoder. 
Additionally, we deploy open-source LLMs using vLLM~\cite{kwon2023efficient} library to ensure efficient inference.
\subsection{Pseudo Code of Genetic Prompt}
We provide the pseudo code of Genetic Prompt in Algorithm~\ref{alg:LLM-GA}.

\subsection{Selected Textual Genes}
\label{sec: selected genes}
The selection of genes for each dataset was tailored to capture the unique characteristics and requirements of their respective domains.
Table~\ref{tab:textual_genes} shows the textual genes selected for the different datasets.
Across all four datasets, common genes such as \textbf{sentence length} and \textbf{voice} were consistently included, reflecting their universal importance in text. 
However, each dataset incorporated specific genes to address its particular focus. 
The Chemprot dataset, dealing with protein-chemical relations, included genes like \textbf{interaction verb} and \textbf{entity proximity} to capture the nuances of biochemical interactions. 
The DDI (Drug-Drug Interaction) dataset emphasized pharmacological aspects with genes such as \textbf{drug mentions} and \textbf{interaction verb}. 
For the more general relation classification tasks, both the SemEval2010 and CoNLL04 datasets shared a set of genes including \textbf{sentence structure}, \textbf{readability}, and \textbf{entity distance}, which are crucial for understanding the contextual relationships between entities.

\subsection{Prompt Template of Genetic Prompt}
\subsubsection{AGNews}

\lstset{
    backgroundcolor=\color[RGB]{245,245,245},
    breaklines=true,
    breakindent=0pt,
    basicstyle=\ttfamily\small,
    frame=trbl,
    frameround = tttt,
}\begin{lstlisting}
You need to generate synthetic data for the Protein Chemical Relation extraction task.

Example 1: {sample1}
Example 2: {sample2}

Your task is to write a sentence about '{class_name}' relation between chemical and protein.
The two entities should be marked with XML-style tags as <chemical> CHEMICAL </chemical> 
and <protein> PROTEIN </protein> respectively in your response.
The sentence should follow the requirements below:
1. The sentence must discuss about the chemical and protein with the relation {label_def}.
2. The '{Gene[0]}' of the sentence should inherit from Example1;
3. The '{Gene[1]}' of the sentence should inherit from Example2;
4. The '{Gene[2]}' of the sentence should inherit from Example1;
5. The '{Gene[3]}' of the sentence should inherit from Example2;
6. The '{Gene[4]}' of the sentence should inherit from Example1;
7. The '{Gene[5]}' of the sentence should inherit from Example2;
8. The '{Gene[6]}' of the sentence and entities must be different from the given 2 examples.
Relation: {class_name}
Text:
\end{lstlisting}

\subsubsection{StackExchange}

\lstset{
    backgroundcolor=\color[RGB]{245,245,245},
    breaklines=true,
    breakindent=0pt,
    basicstyle=\ttfamily\small,
    frame=trbl,
    frameround = tttt,
}\begin{lstlisting}
You need to generate synthetic data for the News Classification task.

Example 1: {sample1}
Example 2: {sample2}

Your task is to write a news article about '{class_name}' category.
The article should follow the requirements below:
1. The article must be a news about {label_def}.
2. The '{Gene[0]}' of the sentence should inherit from Example1;
3. The '{Gene[1]}' of the sentence should inherit from Example2;
4. The '{Gene[2]}' of the sentence should inherit from Example1;
5. The '{Gene[3]}' of the sentence and entities must be different from the given 2 examples.
Category: {class_name}
Text:
\end{lstlisting}
\subsubsection{Chemprot}

\lstset{
    backgroundcolor=\color[RGB]{245,245,245},
    breaklines=true,
    breakindent=0pt,
    basicstyle=\ttfamily\small,
    frame=trbl,
    frameround = tttt,
}\begin{lstlisting}
You need to generate synthetic data for the Stack Exchange question classification task.

Example 1: {sample1}
Example 2: {sample2}

Your task is to write a Stack Exchange question about '{class_name}'.

The '{class_name}' category means {label_def}

The question should follow the requirements below:
1. The '{Gene[0]}' of the sentence should inherit from Example1;
2. The '{Gene[1]}' of the sentence should inherit from Example2;
3. The '{Gene[2]}' of the sentence should inherit from Example1;
4. The '{Gene[3]}' of the sentence and entities must be different from the given 2 examples.
Category: {class_name}
Text:
\end{lstlisting}
\subsubsection{DDI}
\lstset{
    backgroundcolor=\color[RGB]{245,245,245},
    breaklines=true,
    breakindent=0pt,
    basicstyle=\ttfamily\small,
    frame=trbl,
    frameround = tttt,
}\begin{lstlisting}
You need to generate synthetic data for Drug-Drug Interaction extraction task. 

Example 1: {sample1}
Example 2: {sample2}

Your task is to write a sentence about '{class_name}' relation between drug and drug.
The two drugs should be marked with XML-style tags as <drug> DRUG </drug>.
The sentence should follow the requirements below:
1. The sentence must discuss about the drug and drug with the relation {label_def}.
2. The '{Gene[0]}' of the sentence should inherit from Example1;
3. The '{Gene[1]}' of the sentence should inherit from Example2;
4. The '{Gene[2]}' of the sentence should inherit from Example1;
5. The '{Gene[3]}' of the sentence should inherit from Example2;
6. The '{Gene[4]}' of the sentence should inherit from Example1;
7. The '{Gene[5]}' of the sentence should inherit from Example2;
8. The '{Gene[6]}' of the sentence and entities must be different from the given 2 examples.
Relation: {class_name}
Text:
\end{lstlisting}    

\subsubsection{Semeval2010}
\lstset{
    backgroundcolor=\color[RGB]{245,245,245},
    breaklines=true,
    breakindent=0pt,
    basicstyle=\ttfamily\small,
    frame=trbl,
    frameround = tttt,
}\begin{lstlisting}
You need to generate synthetic data for Relation Classification task.

Example 1: {sample1}
Example 2: {sample2}

Your task is to write a sentence about '{class_name}' relation between 2 entities.
The '{class_name}' relation means {label_def}
The two entities should be marked with XML-style tags as <entity> ENTITY </entity>.
The sentence should follow the requirements below:
1. The '{Gene[0]}' of the sentence should inherit from Example1;
2. The '{Gene[1]}' of the sentence should inherit from Example2;
3. The '{Gene[2]}' of the sentence should inherit from Example1;
4. The '{Gene[3]}' of the sentence should inherit from Example2;
5. The '{Gene[4]}' of the sentence should inherit from Example1;
6. The '{Gene[5]}' of the sentence and entities must be different from the given 2 examples.
Relation: {class_name}
Text:
\end{lstlisting}
\subsubsection{Conll04}
\lstset{
    backgroundcolor=\color[RGB]{245,245,245},
    breaklines=true,
    breakindent=0pt,
    basicstyle=\ttfamily\small,
    frame=trbl,
    frameround = tttt,
}\begin{lstlisting}
You need to generate synthetic data for Relation Classification task.

Example 1: {sample1}
Example 2: {sample2}

Your task is to write a sentence about '{class_name}' relation between 2 entities.
The '{class_name}' relation means {label_def}
The two entities should be marked with XML-style tags as <entity> ENTITY </entity>.
The sentence should follow the requirements below:
1. The '{Gene[0]}' of the sentence should inherit from Example1;
2. The '{Gene[1]}' of the sentence should inherit from Example2;
3. The '{Gene[2]}' of the sentence should inherit from Example1;
4. The '{Gene[3]}' of the sentence should inherit from Example2;
5. The '{Gene[4]}' of the sentence should inherit from Example1;
6. The '{Gene[5]}' of the sentence and entities must be different from the given 2 examples.
Relation: {class_name}
Text:
\end{lstlisting}

\subsubsection{SciTLDR}
\lstset{
    backgroundcolor=\color[RGB]{245,245,245},
    breaklines=true,
    breakindent=0pt,
    basicstyle=\ttfamily\small,
    frame=trbl,
    frameround = tttt,
}\begin{lstlisting}
You need to generate synthetic data for the computer science paper abstract summarization task.

Example 1: {sample1}
Example 2: {sample2}

Your task is to write computer science paper abstract and it's summary.
The abstract and summary should follow the requirements below: 
1. The '{Gene[0]}' of the sentence should inherit from Example1;
2. The '{Gene[1]}' of the sentence should inherit from Example2;
3. The '{Gene[2]}' of the sentence should inherit from Example1;
4. The '{Gene[3]}' of the sentence and entities must be different from the given 2 examples.

Output:
\end{lstlisting}
\subsubsection{MeQSum}
\lstset{
    backgroundcolor=\color[RGB]{245,245,245},
    breaklines=true,
    breakindent=0pt,
    basicstyle=\ttfamily\small,
    frame=trbl,
    frameround = tttt,
}\begin{lstlisting}
You need to generate synthetic data for the medical question summarization task.

Example 1: {sample1}
Example 2: {sample2}

Your task is to write a medical question and it's summary.
The abstract and summary should follow the requirements below: 
1. The '{Gene[0]}' of the sentence should inherit from Example1;
2. The '{Gene[1]}' of the sentence should inherit from Example2;
3. The '{Gene[2]}' of the sentence should inherit from Example1;
4. The '{Gene[3]}' of the sentence and entities must be different from the given 2 examples.

Output:
\end{lstlisting}
\section{Definition of Diversity and Distribution Metrics}
\label{sec:definition of diversity}
In this work, we use two metrics to evaluate the diversity and similarity within the datasets: Average Pairwise Similarity (APS) and Vocabulary size. We also use the Central Moment Discrepancy (CMD)~\cite{zellinger2022central} between the synthetic and gold data as the measure of distribution shift.

\subsection{Average Pairwise Similarity (APS)}

The APS measures the overall similarity between samples by averaging the cosine similarity between all embeddings. 
Given embeddings \(e_i\) and \(e_j\) for samples \(i\) and \(j\), APS is defined as:

$$
\text{APS} = \frac{1}{N(N-1)} \sum_{i=1}^{N} \sum_{j=1, j \neq i}^{N} \cos(e_i, e_j)
$$

where \(N\) is the total number of samples and \(\ cos(e_i, e_j)\) represent the cosine similarity between \(e_i\) and \(e_j\).

We also compute Intra-Class and Inter-Class APS:

\paragraph{Intra-Class APS} measures the similarity between embeddings within the same class:

$$
\text{Intra-Class APS} = \frac{1}{|S|} \sum_{(i, j) \in S} \cos(e_i, e_j)
$$

where \(S\) is the set of sample pairs with the same class label, \(label_i = label_j\).

\paragraph{Inter-Class APS} measures the similarity between embeddings from different classes:

$$
\text{Inter-Class APS} = \frac{1}{|D|} \sum_{(i, j) \in D} \cos(e_i, e_j)$$

where \(D\) is the set of sample pairs with different class labels, \(label_i \neq label_j\).





\subsection{Central Moment Discrepancy (CMD)}
Central Moment Discrepancy~\cite{zellinger2022central} measures the difference of probability distributions on a compact interval that considers higher order central moments. 
For our computations, we define the compact interval as \([-1,1]^{N}\) and consider the first five central moments. The CMD between two distributions \(p\) and \(q\), with corresponding samples \(X = (X_1, \dots, X_n)\) from \(p\) and \(Y = (Y_1, \dots, Y_n)\) from \(q\), is calculated as:

\begin{align*}
   CMD(p, q) &= \frac{1}{2} \|\mathbb{E}(X) - \mathbb{E}(Y)\|_2 \\
   &\quad + \sum_{k=2}^{5} \frac{1}{2^k} \|\mathbf{c}_k(X) - \mathbf{c}_k(Y)\|_2
\end{align*}

where \(\mathbb{E}(X)\) and \(\mathbb{E}(Y)\) represent the expectations (means) of the samples, and \(\mathbf{c}_k(X)\) and \(\mathbf{c}_k(Y)\) are the central moments of order \(k\) for \(X\) and \(Y\), respectively. Each term \(\mathbf{c}_k(X)\) is defined as:

\[
\mathbf{c}_k(X) = \left(\mathbb{E} \left( \prod_{i=1}^N (X_i - \mathbb{E}(X_i))^{r_i} \right) \right)
\]

where \(r_1 + \dots + r_N = k\), and \(r_1, \dots, r_N \geq 0\).


\label{sec:appendix Metrics formula}

\section{Case Study}
\label{sec:appendix-case-study}

In this section, we present a case study of the samples generated by different methods. 
Table~\ref{tab:example} provides representative examples from our Gold dataset and synthetic data produced by (SimPrompt and Curated LLM), AttrPrompt, and our proposed Genetic Prompt method. 

From the Table~\ref{tab:example}, it is evident that the samples generated by SimPrompt and Curated LLM tend to use explicit relational markers such as ``cause'', ``from'', ``make'' or ``lead to'' to directly state the relationships between entities. 
While AttrPrompt introduces some variety with interrogative and compound sentences, its overall style remains relatively uniform. 
In contrast, Genetic Prompt’s outputs do not rely on such overt cues. For example, in the sentence ``With the increase in solar activity, satellites experience disruptions in their communications systems.'' the causal relationship is implied through the context and structure rather than being explicitly signaled by words like ``cause'' or ``lead to''.
This implicit expression results in a highly heterogeneous mix of sentence structures, styles, and lengths.

\begin{table*}[htp]
    \centering
    \renewcommand{\arraystretch}{1.2} 
    \setlength{\tabcolsep}{8pt} 
    \begin{tabular}{l p{12cm}} 
        \hline
        \textbf{Method} & \textbf{Example} \\
        \hline
        \multirow{4}{*}{\textbf{Gold}} & 1. The \textbf{burst} has been caused by water hammer \textbf{pressure}. \\
        & 2. He had chest pains and \textbf{headaches} from \textbf{mold} in the bedrooms. \\
        & 3. The \textbf{heavy rainstorm} caused \textbf{severe flooding} in the downtown area, leading to widespread disruptions and evacuations. \\
        \hline
        \textbf{SimPrompt} & 1. The rise in \textbf{global temperatures} is causing the \textbf{melt of polar ice caps}. \\
                
        \multirow{2}{*}{\textbf{And}} & 2. The intense \textbf{heatwave} led to \textbf{crop failures} in several rural communities, \\
        & significantly impacting the livelihoods of local farmers. \\
        
        \textbf{Curated LLM} & 3. \textbf{Smoking} leads to \textbf{cancer} in many cases. \\

        \hline
        \multirow{5}{*}{\textbf{AttrPrompt}} & 1. Increase the budget for the \textbf{marketing} team, as the new \textbf{strategies} will cause a significant impact on sales. \\
        & 2. Is the \textbf{landscape} shaped by \textbf{erosion}? \\
        & 3. The \textbf{storm} was responsible for causing damage, and as a result, the \textbf{power} outage occurred because of it. \\
        \hline
        \multirow{7}{*}{\textbf{Genetic Prompt}} & 1. With the \textbf{increase} in solar \textbf{activity}, satellites experience disruptions in their communications systems. \\
        & 2. The political upheaval resulted in the region a devastating financial setback due to the \textbf{scandal} surrounding the mismanagement of public \textbf{funds} during the last fiscal cycle. \\
        & 3. She observed changes in the plant growth following the application of the \textbf{fertilizer} after the rainfall \textbf{event}. \\
        \hline
    \end{tabular}
    \caption{Text Examples in \textit{Cause-effect} class in Semeval dataset.}
    \label{tab:example}
\end{table*}

\end{document}